\documentclass[10pt,twocolumn,letterpaper]{article}

\usepackage{wacv}
\usepackage{times}
\usepackage{epsfig}
\usepackage{graphicx}
\usepackage{subcaption}
\usepackage{amsmath}
\usepackage{amssymb}
\usepackage{booktabs}
\usepackage{multirow}

\usepackage{dblfloatfix}

\def\wacvPaperID{789} 

\wacvfinalcopy 

\ifwacvfinal
\def\assignedStartPage{1} 
\fi

\ifwacvfinal
\usepackage[breaklinks=true,bookmarks=false]{hyperref}
\else
\usepackage[pagebackref=true,breaklinks=true,colorlinks,bookmarks=false]{hyperref}
\fi

\ifwacvfinal
\else
\fi

\usepackage[numbers]{natbib}
\begin{document}


\title{DORi: Discovering Object Relationship for Moment Localization of a Natural-Language Query in Video}

\author{Cristian Rodriguez-Opazo$^{1,2}$ \qquad Edison Marrese-Taylor$^{3}$ \qquad Basura Fernando$^{4}$\\ \qquad Hongdong Li$^{1,2}$ \qquad Stephen Gould$^{1,2}$ \\
\\
${}^{1}$ Australian National University, \quad
${}^{2}$ Australian Centre for Robotic Vision (ACRV)\\{\tt\small \{cristian.rodriguez, hongdong.li, stephen.gould\}@anu.edu.au}\\
${}^{3}$ Graduate School of Engineering, The University of Tokyo \quad
${}^{4}$ A*AI, A*STAR Singapore\\
{\tt\small emarrese@weblab.t-utokyo.ac.jp} {\tt\small fernando.basura@scei.a-star.edu.sg}
}

\maketitle

\begin{abstract}
    This paper studies the task of temporal moment localization in a long untrimmed video using natural language query.  Given a query sentence, the goal is to determine the start and end of the relevant segment within the video.
    Our key innovation is to learn a video feature embedding through a language-conditioned message-passing algorithm suitable for temporal moment localization which captures the relationships between humans, objects and activities in the video. These relationships are obtained by a spatial sub-graph that contextualizes the scene representation using detected objects and human features conditioned in the language query. Moreover, a temporal sub-graph captures the activities within the video through time. Our method is evaluated on three standard benchmark datasets, and we also introduce YouCookII as a new benchmark for this task. Experiments show our method outperforms state-of-the-art methods on these datasets, confirming the effectiveness of our approach.
\end{abstract}

\section{Introduction}   
\label{sec:intro}
Video analysis using natural language has been drawing increasing attention from the computer vision and natural language communities over the past few years, acknowledging the importance of these two modalities to understand video content. While promising results have been achieved on the tasks of video captioning and video question answering, much work still needs to be done to help identify and trim informative video segments in longer videos and align them with relevant textual descriptions. For this reason, tasks such as automatically recognizing \textit{when} an activity is happening in a video have recently become crucial for video analysis. 

Moreover, as the amount of video data continues to grow, searching for specific visual events in large video collections has become increasingly relevant for search engines. This search engine requirement has helped draw increased attention to the task of activity detection in recent years. This task is especially important, considering that manually annotating videos is laborious and error-prone, even for a small number of videos. In this sense, it is clear that search engines have to retrieve videos not only based on video metadata but that they must also consider the videos' content in order to localize a given query accurately. Applications in practical areas such as video surveillance and robotics~\cite{Liu_ICRA_2018} have also helped bring interest in this task. 

\begin{figure}[t]
    \centering
    \includegraphics[width=0.5\textwidth]{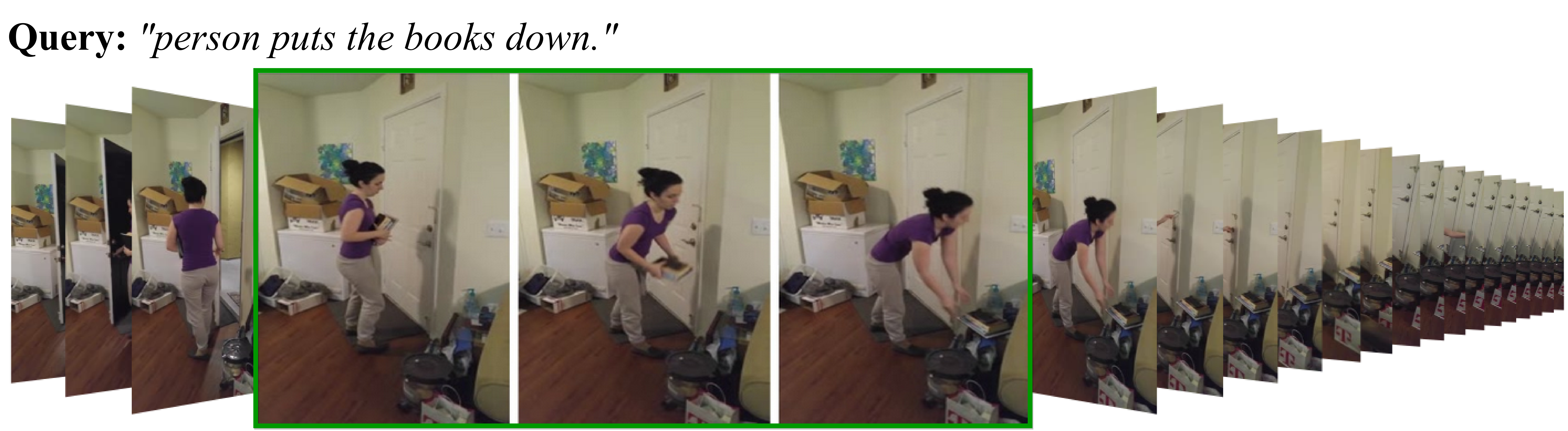}
    \caption{An illustration of temporal localization of a natural language query in an untrimmed video. Given a query and a video the task is to identify the temporal start and end of the sentence in the video.}
    \label{fig:example}
\end{figure}

Regarding the issue mentioned above, in this paper we are specifically interested in the task of temporal sentence localization, in which given an untrimmed video and a natural language query, the goal is to identify the start and end points of the video segment (i.e., moment) that best corresponds to the given query. This task can be seen as a generalization of temporal action localization task \cite{richard2018neuralnetwork,lin_single_2017,escorcia_daps:_2016,Chao2018,GaoYSCN17,xu2019multilevel}.

Many of the existing approaches to the localization problem in vision-and-language, either spatial or temporal, have focused on creating a good multi-modal embedding space and generating proposals based on the given query. In these \textit{propose and rank} approaches, candidate regions are first generated by a particular method and then fed to a classifier to get the probabilities of containing target classes, effectively ranking them. Most recently, approaches that do not rely on proposals to tackle this task have also been proposed, including the work of Ghosh et al. \cite{ghosh2019excl} and Rodriguez et al. \cite{rodriguez2019proposal}.
    
Evidence shows that solving grounded language tasks such as ours often requires reasoning about relationships between objects in the context of the task~\cite{huLanguageConditionedGraphNetworks2019}. For example, the work of Sigurdsson et al. \cite{sigurdsson_ICCV_2017} showed that the performance in action recognition tasks improves by a large margin if we have a perfect object recognition oracle. Moreover, we note that the majority of the queries that are used for this task are related to human actions. Our primary motivation is to reason about the relationship between humans and objects with the activity that they are performing. One can `read a book' or `look at the mobile.' A good way to know what the person is doing is to make use of object clues. 

In light of this, in this paper we propose a mechanism to obtain contextualized activity representations based on a language-conditioned message passing algorithm. As activities are usually the result of the composition of several actions or interactions between a subject and objects \cite{jiang_IJMIR_2013}, our algorithm incorporates both spatial and temporal dependencies. Therefore, modeling the relationship between subjects and objects in a scene and how these change over time, supporting the temporal moment localization task. 

We conduct experiments on four challenging datasets, Charades-STA \cite{Gao_2017_ICCV}, ActivityNet \cite{caba2015activitynet,Krishna_2017_ICCV}, TACoS \cite{tacos} and YouCookII \cite{ZhXuCoCVPR18,ZhLoCoBMVC18}, demonstrating the effectiveness of our proposed method and obtaining state-of-the-art performance. 
Our results highlight the importance of our message-passing algorithm in modeling the relationship between human and object and their interaction to understand the activity, ultimately validating our proposed approach. 
Our approach is the first to incorporate a language-conditioned message-passing algorithm to obtain contextualized activity representations using the objects and subjects to the best of our knowledge. 

\section{Related Work} 
\label{sec:related_work}

Our work is related to the temporal action localization task, which aims to recognize and determine the temporal boundaries of action instances in videos. There is extensive previous work on this task, ranging from models that train existing video feature extractors with a localization loss \cite{Shou_2016_CVPR}, to systems that generally rely on temporal action proposal, as well as more sophisticated models that perform contextual modelling, capturing objects and their interactions \cite{guAVAVideoDataset2018,girdharVideoActionTransformer2019}.

Since action localization is restricted to a pre-defined list of options, Gao et al. \cite{Gao_2017_ICCV} and Hendricks et al. \cite{Hendricks_2017_ICCV} introduced a variation of the task called language-driven temporal moment localization, where the goal is to determine the start and end time of the temporal video segment that best corresponds to a given natural language query. Early approaches for this task, including Liu et al. \cite{liu2018attentive} and Ge et al. \cite{ge2019mac}, were mainly based on generating proposals or candidate clips which could later be ranked. More recently, Chen et al. \cite{chen-etal-2018-temporally}, Chen and Jiang \cite{sap2019}, and Xu et al. \cite{xu2019multilevel}, have worked on reducing the number of proposals by producing query-guided or query-dependent approaches.





Despite their ability to provide coarse control over the video snippets, proposal-based methods suffer from the computationally expensive candidate proposal matching, which has led to the development of methods that can directly output the temporal coordinates of the segment. In this context, Yuan et al. \cite{yuan2018find} first proposed to use a co-attention-based model, and soon after Ghosh et al. \cite{ghosh2019excl} focused directly on predicting the start and end frames using regressions. More recently, Rodriguez et al. \cite{rodriguez2019proposal} used dynamic filters and modeled label uncertainty to further improve performance, while Mun et al. \cite{munLocalGlobalVideoTextInteractions2020} and Zeng et al. \cite{zengDenseRegressionNetwork2020} proposed more sophisticated modality matching strategies. Compared to these works, although our approach is also proposal-free, we differ in the sense that we aim at incorporating specific spatial information that is useful for the localization problem. 

Our work is also related to context modeling in action recognition. In this context, structural-RNN \cite{jainStructuralRNNDeepLearning2016} models a spatio-temporal graph using an RNN mixture that is differentiable, with applications on human motion modeling and human activity detection. While we build on top of a concept similar to this, we inject the language component into the spatio-temporal graph and focus on the task of temporal moment localization of a natural language query. Our method adds the language into the pipeline using an attention mechanism that captures the objects' and subjects' interactions at the language level.

\begin{figure*}[t]
    \centering
    \includegraphics[width=0.85\textwidth]{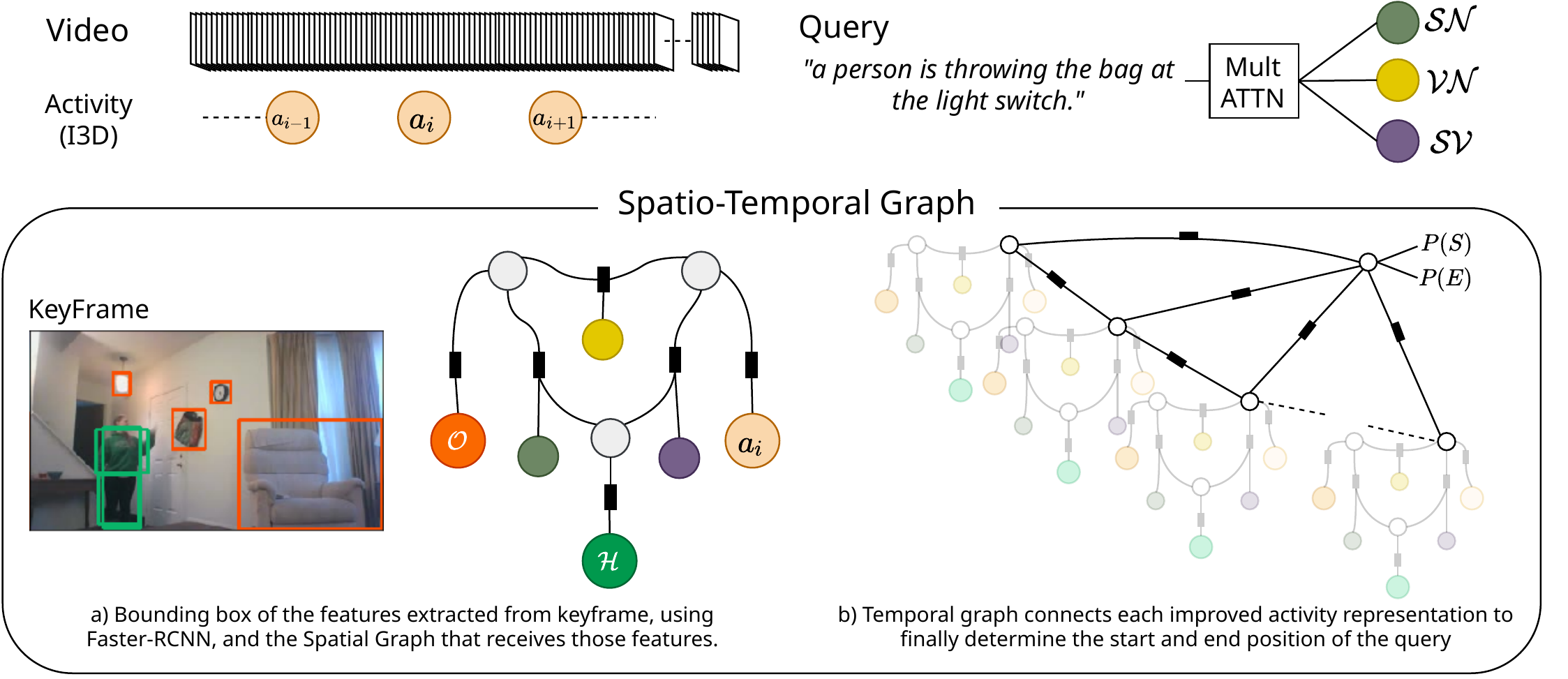}%
    \caption{For each activity feature $a_i$, we create a Spatial graph to find the relationship between object and human nodes conditioned in the query, and thus improve the activity representation to be used by the Temporal graph.}
    \label{fig:method:overview}
    \vspace{-2em}
\end{figure*}
Context modeling has also been recently utilized in other computer vision tasks, such as referring expression comprehension \cite{yuMAttNetModularAttention2018} and VQA \cite{huLanguageConditionedGraphNetworks2019}. In the latter, the authors proposed a Language-Conditioned Graph Network (LCGN) where each node represents an object and is described by a context-aware representation from related objects through iterative message-passing conditioned on the textual input. Our work is fundamentally different from this as our task requires us to model the temporal component in our graph. Moreover, LCGN emphasizes the role of edge representations in the graph, whereas our approach is node-centric as connections between two given node types share the same edges. 

Furthermore, Zeng et al. \cite{zengGraphConvolutionalNetworks2019} used graph convolutions to obtain contextualized representations for action localization, while Zhang et al. \cite{zhang2018man} utilized a graph-structured network to model temporal relationships among different moments and thus obtain contextualized moment representations. Their approach is different from ours as they rely on proposals to perform the task. More recent approaches, such as SLTFNet \cite{jiang_sltfnet_2019}, rely on attention instead of message-passing to deal with the spatio-temporal nature of the moment localization task. 




Finally, Zhang et al. \cite{zhangWhereDoesIt2020} have recently proposed a novel task that requires not only to perform temporal language-driven moment localization but also to locate the objects mentioned in the query spatially. Their approach is similar to ours in the sense that it also utilizes a spatio-temporal graph. However, the textual clues are incorporated after the graph construction rather than being an explicit part of it.

\section{Graph-Based Temporal Moment Localization}
\label{sec:method}

In temporal moment localization, the objective is to find the temporal location of a natural language query $Q$ in an untrimmed video $V$. The video consists of a sequence of frames $V=[ v_t \mid t = 1, \ldots, \ell ]$ and the query is a sequence of words $Q = [ w_j \mid j = 1, \ldots, m ]$ that describes a short moment in the video. We denote the starting and ending times of the moment described by query $Q$ as $t^s$ and $t^e$, respectively. 

We propose a model that explicitly captures the relationship between objects and humans, as well as the activities performed in a video, using a spatio-temporal graph. Concretely, we utilize a language-conditioned message-passing algorithm, which allows us to obtain contextualized activity representations for better moment localization. Let $\mathcal{G} = (\mathcal{V}, \mathcal{E}_S \cup \mathcal{E}_T)$ represent our spatio-temporal graph, where $\mathcal{V}$, $\mathcal{E}_S$ and $\mathcal{E}_T$ are the set of nodes, spatial edges and temporal edges, respectively, as can be seen in Figure \ref{fig:method:overview}.

We factorize our spatio-temporal graph into spatial and temporal sub-graphs, denoted by $\mathcal{G}_{S} = (\mathcal{V}, \mathcal{E}_S)$ and $\mathcal{G}_{T} = (\mathcal{V}, \mathcal{E}_T)$, respectively.

The spatial graph is designed to improve the \emph{activity representations} by exploiting the relationships between objects and humans in a given scene conditioned on an attended language representation for each of this relationships. 
As we know \cite{sigurdsson2017actions}, actions and moments are characterized by complex interactions between humans as well as human-object interactions. Our spatial sub-graph is designed to exploit these spatial relationships specifically. It is iteratively applied through the video (Figure \ref{fig:method:overview}.a.)

On the other hand, the temporal sub-graph is designed to model the relationships between the improved activity representations at different times to more efficiently localize the start and end points of the query in the video (Figure \ref{fig:method:overview}.b.) 

\subsection{Spatial graph} 

Consider the query presented in Figure \ref{fig:method:overview}, ``\textit{a person is throwing the bag at the light switch}''. It describes what action (\textit{verb}) is performed by a \textit{subject} and what \textit{objects} are involved in that action. Our spatial graph is designed to capture the relationships between these visual entities conditioned on the linguistic entities. As such, we decompose our graph into six semantically meaningful nodes, three for representing visual information and three for representing linguistic information. 

\subsubsection{Linguistic Nodes}

We create language nodes to capture essential information in the query related to the visual input: the \textit{subject-verb} relationship node $\mathcal{SV}$ (person-throwing), the \textit{subject-object} relation node $\mathcal{SN}$ (person-bag/light switch) and the \textit{verb-object} relation node $\mathcal{VN}$ (throwing-bag/light switch). 

To obtain representations for each one of the linguistic nodes, we start by encoding each of the words $w_j$ for $j = 1, \ldots, m$ in the query $Q$ using a function $F_w: w \mapsto h$, which maps each word to a semantic embedding vector $h_j \in \mathbb{R}^{d_w}$, where $d_w$ defines the hidden dimension of the word embedding. Specifically, we use GLoVe embeddings \cite{pennington_2014_EMNLP_glove} to obtain the vector representations for each word.

We then initialize three-headed multi-head attention module \cite{vaswani2017attention} using an aggregated, fixed-length query vector $q$. Concretely, we construct this vector using a bi-directional GRU \cite{chung2014empirical} over the word embeddings and mean pooling, which allows us to more accurately capture the global meaning of the input query by first contextualizing each word representation. We project each of the word embeddings using a linear mapping to obtain the key $k$ components of multi-head attention. In the case of the values $v$ we use the contextualized word representations from the GRU. Each head attends these contextualized vectors and returns a re-weighted combination of them, using $\text{softmax}(qk^{\top})v$ and aimed at understanding a specific relation between the visual nodes at the linguistic level.

\subsubsection{Visual Nodes}

As mentioned above, our spatial graph contains three semantically meaningful nodes that represent visual information, specifically an activity node $\mathcal{A}$, a human node $\mathcal{H}$ and an object node $\mathcal{O}$. This setting allows us to share factors for semantically similar observations taken from the video \cite{jainStructuralRNNDeepLearning2016}, which provides several advantages. First, the model can deal with more observations of objects and humans without increasing the number of parameters that need to be learnt. Second, we alleviate the problem of having jittered object detections in videos, specially due to objects appearing and disappearing across frames.

To capture the relationships between activity, human and object observations, we densely connect these nodes within a single video frame. Such relationships are commonly parameterized by factor graphs that convey how a function over the graph factorizes into simpler functions~\cite{kschischang2001factor}. Similarly, we
learn a non-linear mapping function for each of the semantically alike observations that are associated with the same semantic node. In this sense, each semantic node, human $\mathcal{H}$, object $\mathcal{O}$ and activity $\mathcal{A}$, is considered to be a latent representation of the corresponding observation. Let us take as an example the case of the object node $\mathcal{O}$, where we observe a \textit{table} in the video, represented by a feature vector $x$, obtained directly from the object detector. In this case, we use a function $\Psi_{\scriptscriptstyle\mathcal{O}} \doteq \tanh ( W_{\scriptscriptstyle\mathcal{O}} x + b_{\scriptscriptstyle\mathcal{O}} )$. Similar mapping functions (with different parameters), namely $\Psi_{\scriptscriptstyle\mathcal{H}}$ and  $\Psi_{\scriptscriptstyle\mathcal{A}}$ are defined for the other semantic nodes.

\textbf{Activity node}: We use a video encoder that generates a video representation summarizing spatio-temporal patterns directly from the raw input frames. Concretely, let $F_V$ be our video encoding function that maps a video into a sequence of vectors $[ a_i \in \mathbb{R}^{d_v} \mid i = 1, \ldots, t]$.
These features capture high-level visual semantics in the video. 
Note that length of the video, $\ell = |V|$, and the number of output features, $t = |F_V(V)|$, are different due to temporal striding.
Specifically, in this work we model $F_V$ using I3D \cite{carreira2017quo}. This method inflates the 2D filters of a well-known convolutional neural network, e.g., Inception \cite{szegedy2015going,ioffe2015batch} or ResNet \cite{He_2016_CVPR} for image classification to obtain 3D filters. 

\textbf{Human and object nodes}: Activity representations are obtained using small clips of frames. This means that there may be a set of many frames from where to extract spatial information that is semantically relevant for each node.
Utilizing every frame is computationally expensive and given the piece-wise smooth nature of video, this could also prove to be redundant. As such, in this work we propose to utilize key-frames associated to each activity representation to extract observations for human and object nodes. Since many frames in a video are blurry due to various reasons, e.g., the natural movement of the objects and the camera  motion, we select the sharpest key-frame in the subset of frames. Here we use the Laplace variance algorithm \cite{pech2000diatom}, which is a well-known approach for measuring the sharpness of an image.

While our method is agnostic to the choice of object detector, in this work we use Faster RCNN~\cite{ren_NIPS_2015_faster,anderson2018bottom} for the detection and spatial representation of the objects in all key-frames. Our Faster RCNN detector is trained on the Visual Genome \cite{krishnavisualgenome} dataset, which consists of 1,600 object categories. These categories are manually assigned to either the human and object nodes depending on the type of object. The human node receives the set of features $H = \{h_1,...,h_K\}$ corresponding to the categories associated to human body parts, clothes and subjects, while the object node receives the set of features that are not associated to human labels with that $O = \{o_1,...,o_J\}$. This label-based categorization is based on a manual analysis of the label names supported by the Faster RCNN detector. In this way, when taking the predicted labels for each object we can use our categorization to re-label them as human or object and thus assign each instance to their corresponding visual node.




\subsubsection{Language-conditioned message-passing}


We argue that the setting of the scene contains important clues to improve the representation of a given activity. Examples of these clues are human clothes, objects that are present in the scene as well as their appearance. To the best of our knowledge, previous work on moment localization has not utilized this information. 
Therefore, we propose to obtain an activity representation suitable for the moment localization task, by capturing object, human and activity relationships. Concretely, we use a mean-field like approximation of the message-passing algorithm to capture such relationships. The messages sent between nodes are conditioned on the natural language query. We propose to use this approximation instead of the standard message-passing algorithm due to high demand on memory and compute, specially to process all the key-frames in a given video. The messages are iteratively sent a total of $N$ times, which is a hyperparameter of our model. In the equations below, index $n=1,...,N$ denotes the iteration step for each of the nodes. Notice that in the rest of this subsection, we drop the temporal index $i$ in the activity feature $a$ since the message-passing is done for each of the activity features independently.

First, we capture the relationship between the visual observations of the nodes human $\mathcal{H}$, object $\mathcal{O}$ and activity $\mathcal{A}$ with the corresponding language nodes $\mathcal{SN}, \mathcal{SV}$ and $\mathcal{VN}$ that connect the semantic meaning of the visual nodes, using a linear mapping function $f$ specific for each node. For instance, in the case of the object observations, the mapping functions $f$ have the following shape.
{\fontsize{9.5}{12}
\begin{align}
f_{\scriptscriptstyle\mathcal{SN,O}}(\mathcal{SN},o^{j,n}) &= W_{sno} [\mathcal{SN};o^{j,n}] + b_{sno}=\Phi_{\scriptscriptstyle\mathcal{SN,O}}^{j,n} \\
f_{\scriptscriptstyle\mathcal{VN,O}}(\mathcal{VN},o^{j,n}) &= W_{vno} [\mathcal{VN};o^{j,n}] + b_{vno}=\Phi_{\scriptscriptstyle\mathcal{VN,O}}^{j,n}
\end{align}
}%
where $j$ is the j-th object observation in the object node $\mathcal{O}$. Similarly, we have specific mapping functions {\fontsize{9}{12}$f_{\scriptscriptstyle\mathcal{SV,A}}(\mathcal{SV},a^{n}) = \Phi_{\scriptscriptstyle\mathcal{SV,A}}^{n}$, $f_{\scriptscriptstyle\mathcal{VN,A}}(\mathcal{VN},a^{n})=\Phi_{\scriptscriptstyle\mathcal{VN,A}}^n$, $f_{\scriptscriptstyle\mathcal{SN,H}}(\mathcal{SN},h^{k,n})=\Phi_{\scriptscriptstyle\mathcal{SN,A}}^{k,n}$, and $f_{\scriptscriptstyle\mathcal{SV,H}}(\mathcal{SV},h^{k,n})=\Phi_{\scriptscriptstyle\mathcal{SV ,A}}^{k,n}$} for the activity and human observations, where $k$ is the k-th human observation.

For clarity, we explain the message-passing algorithm again using the object node as an example. The object node $\mathcal{O}$ receives messages from the human $\mathcal{H}$ and the activity $\mathcal{A}$ nodes. 
The message from the human node is constructed using a linear mapping function that receives as an input the concatenation of the object-query relationship $\Phi_{\scriptscriptstyle\mathcal{SN,O}}^{j,n}$ and the aggregation of all the human-query relationships $\sum_k \Phi_{\scriptscriptstyle\mathcal{SN,H}}^{k,n}$. A similar process is done for the message received from the activity observations, as can be seen in Equation \ref{object:message1}, below.
\begin{align}
    \Psi^{j,n}_{\scriptscriptstyle\mathcal{H,SN,O}} &= f_{\scriptscriptstyle\mathcal{H,SN,O}}(\Phi_{\scriptscriptstyle\mathcal{SN,O}}^{j,n},\textstyle \sum_{k=1}^K\Phi_{\scriptscriptstyle\mathcal{SN,H}}^{k,n}) \\
    \Psi^{j,n}_{\scriptscriptstyle\mathcal{A,VN,O}} &= f_{\scriptscriptstyle\mathcal{A,VN,O}}(\Phi_{\scriptscriptstyle\mathcal{VN,O}}^{j,n},\Phi_{\scriptscriptstyle\mathcal{VN,A}}^{n})
    \label{object:message1}
\end{align}
\vspace{-0.5 em}
\vspace{-1em}
{\fontsize{10.85}{12}
    \begin{align}
        o^{j,n+1} &= \sigma(m_o(\Psi^{j,n}_{\scriptscriptstyle\mathcal{H,SN,O}} \odot \Psi^{j,n}_{\scriptscriptstyle\mathcal{A,VN,O}}) \odot o^{j,0})
        \label{object:message2}
    \end{align}
}
Finally, the new representation of the object observation is computed using Equation \ref{object:message2}, where $o^{j,0}$ is the initial object representation, $\sigma$ is an activation function, $\odot$ is the Hadamard product and $m_o$ is a linear function with a bias that constructs the message for the object $o^j$. A similar process is applied for each observation, as can be seen in Equations \ref{other:message1a} to \ref{other:message1b} below, where we create the message for each edge and Equations \ref{other:message2a} to \ref{other:message2b} and show how these messages are later used to contextualize the features. Note that the parameters learnt for each specific case are shared. For instance, parameters for $f_{\scriptscriptstyle\mathcal{A,SV,H}}$ and $f_{\scriptscriptstyle\mathcal{H,SV,A}}$ are the same.
\begin{align}
\Psi^{n}_{\scriptscriptstyle\mathcal{H,SV,A}} &= f_{\scriptscriptstyle\mathcal{H,SV,A}}(\Phi_{\scriptscriptstyle\mathcal{SV,A}}^{n},\textstyle \sum_{k=1}^K \Phi_{\scriptscriptstyle\mathcal{SV,H}}^{k,n}) 
\label{other:message1a}\\ 
\Psi^{n}_{\scriptscriptstyle\mathcal{O,VN,A}} &= f_{\scriptscriptstyle\mathcal{O,VN,A}}(\Phi_{\scriptscriptstyle\mathcal{VN,A}}^{n},\textstyle \sum_{j=1}^{J} \Phi_{\scriptscriptstyle\mathcal{VN,O}}^{j,n}) \\
\Psi^{k,n}_{\scriptscriptstyle\mathcal{O,SN,H}} &= f_{\scriptscriptstyle\mathcal{O,SN,H}}(\Phi_{\scriptscriptstyle\mathcal{SN,H}}^{k,n},\textstyle \sum_{j=1}^{J} \Phi_{\scriptscriptstyle\mathcal{SN,O}}^{j,n}) \\
\Psi^{k,n}_{\scriptscriptstyle\mathcal{A,SV,H}} &= f_{\scriptscriptstyle\mathcal{A,SV,H}}(\Phi_{\scriptscriptstyle\mathcal{SV,H}}^{k,n},\Phi_{\scriptscriptstyle\mathcal{SV,A}}^{n})
\label{other:message1b}
\end{align}
\vspace{-2.5em}
{\fontsize{10.25}{12}
\begin{align}
        a^{n+1} &= \sigma(m_a (\Psi^{n}_{\scriptscriptstyle\mathcal{H,SV,A}} \odot \Psi^{n}_{\scriptscriptstyle\mathcal{O,VN,A}}) \odot a^{0}) 
        \label{other:message2a}\\
        h^{k,n+1} &= \sigma(m_h (\Psi^{k,n}_{\scriptscriptstyle\mathcal{O,SN,H}} \odot \Psi^{k,n}_{\scriptscriptstyle\mathcal{A,SV,H}}) \odot h^{k,0}) 
        \label{other:message2b}
\end{align}
}
\subsection{Temporal graph}
The temporal graph is responsible for predicting the starting and ending points of the moment in the video. It uses the previously computed activity representations $a^{i,N} \text{ for } i=1, \ldots, t$ where $N$ is the final iteration in the message passing. 
The temporal graph is implemented using a 2-layer bi-directional GRU \cite{cho-etal-2014-learning} which receives as input the improved activity representation, and it is designed to contextualize the temporal relationship between the activity features. To obtain a probability distribution for the start and end predicted positions, we utilize two different fully connected layers to produce scores associated to the probabilities of each output of the GRU being the start/end of the location. Then, we take the softmax of these scores and thus obtain vectors $ \hat{\boldsymbol {\tau}}^s, \hat{\boldsymbol{\tau}}^e \in \mathbb{R}^{T}$ containing a categorical probability distribution. Even though we do not constrain the starting and ending points to follow the right order in time, this does not result in any difficulties in practice.

\section{Training}

Our method is trained end-to-end on a dataset consisting of annotated tuples $(V, Q, t^s, t^e)$. Note that each video $V$ may include more than one moment and may therefore appear in multiple tuples. We treat each training sample independently. Given a new video and sentence tuple $(V_r, Q_r)$, our model predicts the most likely temporal localization of the moment described by $Q_r$ in terms of its start and end positions, $t_r^{s\star}$ and $t_{r}^{e\star}$, in the video. We use the Kullback-Leibler divergence and an spatial loss proposed by Rodriguez et al. \cite{rodriguez2019proposal}. We explain this in more detail in the supplemental material. 
Given the predicted/ground truth starting/ending  times of the moment, we use the following loss function during training:
\begin{equation}
    L_{\text{KL}} = \displaystyle D_{\text{KL}}(\hat{\boldsymbol{\tau}}^s \parallel \boldsymbol {\tau}^s) + \displaystyle D_{\text{KL}}(\hat{\boldsymbol {\tau}}^e \parallel \boldsymbol {\tau}^e)
    \label{eq:kl_div}
\end{equation}
where $D_{\text{KL}}$ is the Kullback-Leibler divergence. 
Moreover, inspired by Rodriguez et al. \cite{rodriguez2019proposal}, we use a spatial loss that aims to create activity features that are good at identifying where the action is occurring. This loss, equation \ref{eq:spatial}, receives as input $\textbf{y} = \text{softmax}(g (\textbf{a}))$ where $\textbf{a}$ is the matrix that results by concatenating the improved activity representations over time, and $g$ is a linear mapping that gives us a score for each activity representation. We apply a softmax function over these and our loss penalizes if this normalized score is large for those features associated to positions that lie outside the temporal location of the query. 
\begin{equation}
    L_{\text{spatial}} = - \sum_{i=1}^t (1- \delta_{\tau^s \leq i \leq \tau^e}) \log(1-y^i) 
    \label{eq:spatial}
\end{equation}
where $\delta$ is the Kronecker delta.
The final loss for training our method is the sum of the two individual losses defined previously setting $\mathcal{L} =    L_{\text{KL}} + L_{\text{spatial}}$. 
During inference, we predict the starting and ending positions using the most likely locations given by the estimated distributions, using $\hat{\tau}^s = \text{argmax}(\hat{\boldsymbol{\tau}}^s)$ and $\hat{\tau}^e = \text{argmax}(\hat{\boldsymbol{\tau}}^e)$. Since values correspond to positions in the feature domain of the video, so we convert them back to time positions.

\section{Experiments and Results}
\label{sec:experiments}



To evaluate our proposed approach we work with three widely utilized and challenging datasets, namely Charades-STA \cite{Gao_2017_ICCV}, ActivityNet Caption \cite{caba2015activitynet,Krishna_2017_ICCV} and TACoS \cite{tacos}. In addition to these, we also consider the YouCookII dataset \cite{ZhXuCoCVPR18,ZhLoCoBMVC18}. This decision is motivated by its activity-centric nature as YoucookII is built upon instructional videos making it an excellent candidate to evaluate our proposals. 

\noindent
\textbf{Charades-STA}: Collected from the Charades dataset \cite{sigurdsson2016hollywood} by adding sentences that describe actions in the videos, it consists on a total of 13,898 pairs of queries and temporal locations. We use the predefined train and test splits \cite{Gao_2017_ICCV}. Videos are 31 seconds long on average, with 2.4 moments on average, each being 8.2 seconds long on average.

\noindent
\textbf{ActivityNet Captions}: Introduced by \cite{Krishna_2017_ICCV} this dataset, which was originally constructed for dense video captioning, consists of 20k YouTube videos with an average length of 120 seconds. The videos contain 3.65 temporally sentence descriptions on average, where the average length of the descriptions is 13.48 words. Following previous work, we report the performance of our approach on the two existing validation sets combined \cite{munLocalGlobalVideoTextInteractions2020}.

\noindent
\textbf{MPII TACoS}: Built on top of the MPII Compositive dataset, it consists of videos of cooking activities with detailed temporally-aligned text descriptions. There are 18,818 pairs of sentence and video clips in total, with the average video length being 5 minutes. We use the same splits as \cite{Gao_2017_ICCV}, consisting of 50\% for training, 25\% for validation and 25\% for testing.

\noindent
\textbf{YouCookII}: consists on 2,000 long untrimmed videos from 89 cooking recipes obtained from YouTube by \cite{ZhXuCoCVPR18}. Each step for cooking these dishes was annotated with temporal boundaries and aligned with the corresponding section of the recipe. Similarly to TACoS, the average video length is 5.26 minutes. 

\subsection{Implementation Details}

We first pre-process the videos by extracting features of size $1024$ using I3D feaures with average pooling, taking as input the raw frames of dimension $256 \times 256$, at 25fps. We use the pre-trained model trained on Kinetics for TACoS, ActivityNet and YouCookII released by \cite{carreira2017quo}. For Charades-STA, we use the pre-trained model trained on Charades. We extract the top 15 objects detected in terms of confidence for each of the key-frames using Faster-RCNN. 


All of our models are trained in an end-to-end fashion using ADAM \cite{kingma_adam} with a learning rate of $10^{-4}$ and weight decay $10^{-3}$. As mentioned earlier, our temporal graph is modeled using a two-layer BiGRU. We use a hidden size of $256$ and to prevent over-fitting we add a dropout of $0.5$ between the two layers.

\subsection{Evaluation}

We evaluate our model using two widely used metrics proposed by \cite{Gao_2017_ICCV}. Firstly, we measure recall at various thresholds of the temporal Intersection over Union ($R@\alpha$) obtaining the percentage of predictions that have tIoU with ground truth larger than certain $\alpha$, with threshold values $0.3$, $0.5$ and $0.7$. In addition to that, we also compute and report mean or averaged tIoU (mIoU).


\subsection{Ablation Study}

To show the effectiveness of our proposals we perform several ablation studies each aimed at assessing the contribution of different components of our model. All of our ablative experiments are based on a segment of the training split of the Charades-STA dataset. As mIoU provides a more comprehensive evaluation of the performance of our model we utilized this metric to select the best model configuration for the rest of the experiments in this paper.

Since feed-forwarding through our proposed spatial graph is an iterative process, we first studied the impact on performance of the number of iterations ($N$) utilized in the message-passing algorithm of our full model. As shown in Table \ref{tab:res:n}, we experimented setting $N$ to a minimum value of 0 (where nodes are not updated at all) up to a maximum number of 4 iterations. As expected, performance tends to improve with larger values of $N$, with a saturation point at $N=3$. Based on these results, all of our models in the rest of this paper are trained utilizing three iterations.



\begin{table}[t!]
    \centering
    \caption{Performance when using a different number of iterations ($N$) for the message-passing algorithm, on a subsection of the training split of Charades-STA.}
    \begin{tabular}{c cccc c}
        \toprule
        \bf $N$ & R@0.3 & R@0.5 & R@0.7 & R@0.9 & mIoU \\
        \midrule
        0 & 47.46 & 22.88 & 14.38 & 6.00  & 33.67 \\
        1 & 73.21 & 55.32 & 36.02 & 11.48 & 52.11 \\
        2 & 79.01 & 67.16 & 48.71 & 17.97 & 59.30 \\
        3 & \textbf{79.25} & \textbf{68.41} & \textbf{50.56} & \textbf{19.14} & \textbf{60.29} \\
        4 & 70.99 & 60.31 & 44.16 & 17.32 & 54.01 \\
        \bottomrule
    \end{tabular}
    \vspace{-1em}
    \label{tab:res:n}
\end{table}

\begin{table}[t!]
    \centering
    \caption{Results of our ablation studies, performed on a section of the training split of Charades-STA.}
    \scalebox{0.82}{
    \begin{tabular}{lcccc c}
        \toprule
        \bf Model & R@0.3 & R@0.5 & R@0.7 & R@0.9 & mIoU \\
        \midrule
        (1) No Graph & 44.32 & 13.46 & 7.66 & 2.50 & 31.09 \\
        (2) No Node Types & 74.78 & 61.24 & 43.35 & 14.59 & 55.18 \\
        (3) No $\mathcal{H}$ Node & 75.46 & 60.60 & 43.31 & 15.51 & 55.32 \\
        (4) No $\mathcal{O}$ Node & 75.66 & 61.28 & 44.08 & 15.39 & 56.13 \\
        (5) No LA & 76.79 & 66.32 & 49.92 & 20.87 & 58.93\\
        (6) No $L_{\text{spatial}}$ & 76.79 & 66.60 & \textbf{52.54} & \textbf{23.57} & 59.95 \\
        \midrule
        Full Model & \textbf{79.25} & \textbf{68.41} & 50.56 & 19.14 & \textbf{60.29} \\ 
        \bottomrule
    \end{tabular}}
    \vspace{-1em}
    \label{tab:res:ablations}
\end{table}


Table \ref{tab:res:ablations} summarizes the results of our ablation studies, which include: (1) Concatenating the mean-pooling of the features extracted by Faster RCNN directly with the activity representation, therefore eliminating the human and object nodes (No Graph) to assess the relevance of our graph in using the spatial information. (2) Evaluating the importance of distinguishing between human versus object features by testing how our model performs when assigning all the detected features to one spatial node (No Node Types). In (3) and (4) we remove the use of human (No $\mathcal{H}$) and object (No $\mathcal{O}$) spatial information, respectively. (5) Assessing the contribution of the linguistic nodes (No LA) by modifying our graph so that it only contains a single textual node connected to the rest of the graph in a way analogous to our full model. (6) Testing the importance of the spatial loss $L_{\text{spatial}}$ which encourages our model to focus on the features within the segment of interest. 
As can be seen, the importance of each one of our studied components is validated as ablations always result in consistent performance drops in terms of both mIoU as well as tIoU at the majority of $\alpha$ thresholds. For details about our ablated models please check Sections \ref{appe:ablated:nonodetype} and \ref{appe:ablated:nolanguageatte} in the Supplementary Material. 


\subsection{Comparison with the state-of-the-art}

We start by presenting our results on the YouCookII dataset, which we introduce as a baseline for this task since so far it has not been considered by previous work. On this dataset we consider a random baseline that simply selects an arbitrary video segment as the moment for each example, and also used the official implementation of TMLGA \cite{rodriguez2019proposal} released by its authors as an additional baseline since it is a direct alternative to our approach also being proposal-free. Table \ref{tab:youcookii:res} summarizes our obtained results, where it can be seen that DORi is able to outperform both the random baseline and TMLGA by a large margin, specially on the lower $\alpha$ bands.

\begin{table}[t]
    \centering
    \caption{Performance comparison on YouCookII for different tIoU $\alpha$ levels.}
    \begin{tabular}{l c c c c}
        \toprule
        \textbf{Method} & R@0.3 & R@0.5 & R@0.7 & mIoU \\
        \midrule
        Random & 4.84 & 1.72 & 0.60 & - \\ 
        TMLGA & 33.48 & 20.65 & 10.94 & 23.07 \\
        \midrule
        DORi & \textbf{43.73} & \textbf{29.93} & \textbf{17.61} & \textbf{30.43} \\
        \bottomrule
    \end{tabular}
    \vspace{-1em}
    \label{tab:youcookii:res}
\end{table}

\begin{table*}[t]
    \centering
    \caption{Performance comparison of our approach with existing methods for different tIoU $\alpha$ levels. Values are reported on the validation split of Charades-STA and ActivityNet Captions, and test splits for the TACoS datasets. $\dag$ Results for ABLR are as reported by \cite{sap2019}.}
    \scalebox{0.90}{
    \begin{tabular}{l c c c c c c c c c c c c}
        \toprule
        \multirow{2}{*}{\textbf{Method}} &\multicolumn{4}{c}{Charades-STA} & \multicolumn{4}{c}{ActivityNet} & \multicolumn{4}{c}{TACoS} \\
        \cmidrule{2-13}
        & R@0.3 & R@0.5 & R@0.7 & mIoU & R@0.3 & R@0.5 & R@0.7 & mIoU & R@0.3 & R@0.5 & R@0.7 & mIoU\\
        \midrule
        Random & - & 8.51 & 3.03 & - & 5.60 & 2.50 & 0.80 & & 1.81 & 0.83 & & - \\
        CTRL & - & 21.42 & 7.15 & - & 28.70 & 14.00 & - & 20.54 & 18.90 & 13.30 & - & - \\
        ABLR $\dag$ & - & 24.36 & 9.00 & - & 55.67 & 36.79 & - & 36.99 & 18.90 & 9.30 & - & - \\
        TripNet & 51.33 & 36.61 & 14.50 & - & 48.42 & 32.19 & 13.93 & - & 23.95 & 19.17 & 9.52 & - \\
        CBP & 50.19 & 36.80 & 18.87 & 35.74 & 54.30 & 35.76 & 17.80 & 36.85 & 27.31 & 24.79 & 19.10 & 21.59 \\
        MAN   & - & 46.53 & 22.72 & - & - & - & - & - & - & - & - & - \\
        EXCL  & 65.10 & 44.10 & 22.60 & - & - & - & - & - & \textbf{44.20} & 28.00 & 14.60 & - \\
        TMLGA & 67.53 & 52.02 & 33.74 & 48.22 & 51.28 & 33.04 & 19.26 & 37.78 & 24.54 & 21.65 & 16.46 & 22.06 \\ 
        LGVTI & \textbf{72.96} & 59.46 & 35.48 & 51.38 & \textbf{58.52} & \textbf{41.51} & 23.07 & 41.13 & - & - & - & - \\
        \midrule
        DORi & 72.72 & \textbf{59.65} & \textbf{40.56} & \textbf{53.28} & 57.89 & 41.49 & \textbf{26.41} & \textbf{42.78} & 31.80 & \textbf{28.69} & \textbf{24.91} & \textbf{26.42}\\ 
        \bottomrule
    \end{tabular}
    }
    \label{tab:csta:res}
\end{table*}

\begin{figure*}[h]
    \centering
    \includegraphics[width=0.92\textwidth]{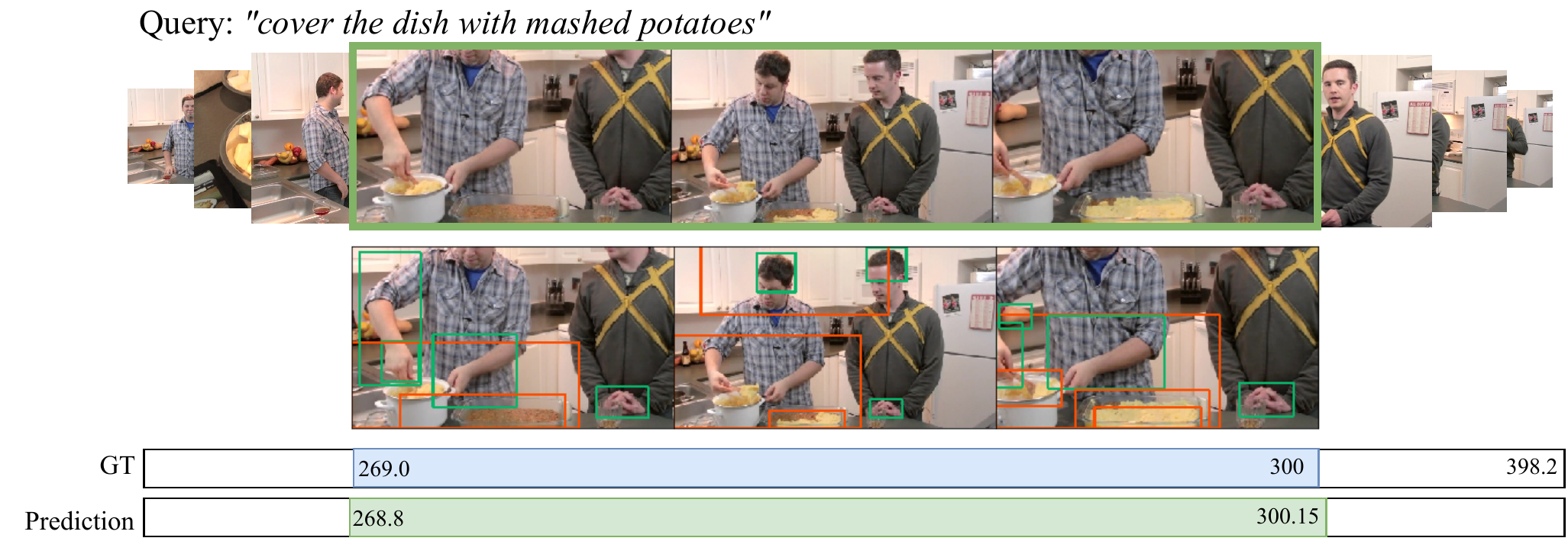}
    \caption{Visualization of a success case of our method in the YouCookII dataset. The second row shows the observations associated to the Human node (green) and Object node (orange).}
    \label{fig:success:youcookii}
    \vspace{-1.2em}
\end{figure*}
Regarding benchmark datasets, we compare the performance of our proposed approach against several prior work selected from the literature. We consider a broad selection of models based on different approaches, specifically proposal-based techniques including CTRL \cite{Gao_2017_ICCV}, SAP \cite{sap2019}, MAN \cite{zhang2018man} and CBP \cite{wang2019temporally}, as well as TripNet \cite{hahn2019tripping}, a method based on reinforcement learning. In addition to that, we also compare our approach to more recent methods that do not rely on proposals, including ABLR \cite{yuan2018find}, ExCL \cite{ghosh2019excl}, TMLGA \cite{rodriguez2019proposal} and LGVTI \cite{munLocalGlobalVideoTextInteractions2020}, as well as our random baseline.

Table \ref{tab:csta:res} summarizes our results on Charades-STA, ActivityNet Captions and TACoS, while also comparing the obtained performance to relevant prior work. It is possible to see that our method is able to outperform previous work by a consistent margin, specially for the $\alpha=0.7$ band and also in terms of the mean tIoU (mIoU). Comparing results across these datasets, we also see that the performance of all models drops substantially on ActivityNet Captions and TACoS, compared to Charades-STA. We think this is mainly due to the nature of the dataset, which contains a considerable amount of video moments that only span a few seconds. 

\begin{table}[t!]
    \centering
    \footnotesize
    \caption{Performance of our method with VGG-16 features, We compared to relevant prior work that uses the same type of features.}
    \begin{tabular}{ l cccc c}
        \toprule
        \bf Model & R@0.3 & R@0.5 & R@0.7 & R@0.9 & mIoU \\
        \midrule
        SAP & - & 27.42 & 13.36 & - & - \\
        MAN & - & 41.24 & 20.54 & - & - \\
        \midrule
        DORi & 61.83 & 43.47 & 26.37 & 7.63 & 42.52 \\
        \bottomrule
    \end{tabular}
    \vspace{-1em}
    \label{tab:res:i3dvsvgg}
\end{table}

Finally, we also study the effect of using different a pre-trained model to obtain activity representations in our proposed approach. Concretely, we test the performance of our model using VGG-16 features instead of I3D on Charades-STA. Table \ref{tab:res:i3dvsvgg} summarizes our obtained results and compares them to prior work also utilizing these features. As can be seen, although VGG features provide lower performance than I3D in our experiments, therefore experimentally validating our choice, our model is still able to outperform existing approaches also using these features by a large margin, showing the superiority of our proposed approach. 

\subsection{Qualitative results}

Figure \ref{fig:success:youcookii} presents a success case of our method on YouCookII dataset. The visualization is presenting a subsample of the key-frames inside of the prediction with their corresponding spatial observations, with green observations associated with the human node $\mathcal{H}$ and orange to the object node $\mathcal{O}$. Moreover, each visualization is presenting the ground-truth localization and predicted localization of the given query. As shown in Figure \ref{fig:success:youcookii}, given the query ``cover the dish with mashed potatoes'', our method could localize the moment at a tIoU of 98.88\%. The most relevant features extracted by Faster-RCNN to localize the query are {\em 'arm', 'bowl', 'cake', 'hand', 'kitchen', 'man', 'mug', 'spoon', 'stove', 'tray'}. Additional qualitative examples of success and failure cases of our method are included in Section \ref{appe:examples} of the Supplementary Material.


\section{Conclusion}

We have presented a novel approach to temporal moment localization in video. Our approach consists of a spatial-temporal graph for capturing the relationships between detected humans, objects and activities over time. Conditioned on a natural language query, we proposed a message-passing algorithm that propagates information across the graph to ultimately infer the arbitrarily long segment in the video most likely described by the query. Using our approach we are able to achieve state-of-the-art results on several benchmark datasets.

{\small
\bibliographystyle{ieee_fullname}
\bibliography{egbib}

\begin{thebibliography}{10}\itemsep=-1pt

\bibitem{anderson2018bottom}
Peter Anderson, Xiaodong He, Chris Buehler, Damien Teney, Mark Johnson, Stephen
  Gould, and Lei Zhang.
\newblock Bottom-up and top-down attention for image captioning and visual
  question answering.
\newblock In {\em Proceedings of the IEEE conference on computer vision and
  pattern recognition}, pages 6077--6086, 2018.

\bibitem{caba2015activitynet}
Fabian Caba~Heilbron, Victor Escorcia, Bernard Ghanem, and Juan Carlos~Niebles.
\newblock Activitynet: A large-scale video benchmark for human activity
  understanding.
\newblock In {\em Proceedings of the IEEE Conference on Computer Vision and
  Pattern Recognition}, pages 961--970, 2015.

\bibitem{carreira2017quo}
Joao Carreira and Andrew Zisserman.
\newblock Quo vadis, action recognition? a new model and the kinetics dataset.
\newblock In {\em CVPR}, 2017.

\bibitem{Chao2018}
Yu{-}Wei Chao, Sudheendra Vijayanarasimhan, Bryan Seybold, David~A. Ross, Jia
  Deng, and Rahul Sukthankar.
\newblock Rethinking the faster {R-CNN} architecture for temporal action
  localization.
\newblock {\em CVPR}, 2018.

\bibitem{chen-etal-2018-temporally}
Jingyuan Chen, Xinpeng Chen, Lin Ma, Zequn Jie, and Tat-Seng Chua.
\newblock Temporally grounding natural sentence in video.
\newblock In {\em Proceedings of the 2018 Conference on Empirical Methods in
  Natural Language Processing}, pages 162--171, Brussels, Belgium, 2018.
  Association for Computational Linguistics.

\bibitem{sap2019}
Shaoxiang Chen and Yu-Gang Jiang.
\newblock Semantic proposal for activity localizaiton in videos via sentence
  query.
\newblock {\em AAAI}, 2019.

\bibitem{cho-etal-2014-learning}
Kyunghyun Cho, Bart van Merri{\"e}nboer, Caglar Gulcehre, Dzmitry Bahdanau,
  Fethi Bougares, Holger Schwenk, and Yoshua Bengio.
\newblock Learning phrase representations using {RNN} encoder{--}decoder for
  statistical machine translation.
\newblock In {\em Proceedings of the 2014 Conference on Empirical Methods in
  Natural Language Processing ({EMNLP})}, pages 1724--1734, Doha, Qatar, Oct.
  2014. Association for Computational Linguistics.

\bibitem{chung2014empirical}
Junyoung Chung, Caglar Gulcehre, KyungHyun Cho, and Yoshua Bengio.
\newblock Empirical evaluation of gated recurrent neural networks on sequence
  modeling.
\newblock {\em arXiv preprint arXiv:1412.3555}, 2014.

\bibitem{escorcia_daps:_2016}
Victor Escorcia, Fabian Caba~Heilbron, Juan~Carlos Niebles, and Bernard Ghanem.
\newblock {DAPs}: {Deep} {Action} {Proposals} for {Action} {Understanding}.
\newblock In Bastian Leibe, Jiri Matas, Nicu Sebe, and Max Welling, editors,
  {\em Computer {Vision} – {ECCV} 2016}, Lecture {Notes} in {Computer}
  {Science}, pages 768--784. Springer International Publishing, 2016.

\bibitem{Gao_2017_ICCV}
Jiyang Gao, Chen Sun, Zhenheng Yang, and Ram Nevatia.
\newblock Tall: Temporal activity localization via language query.
\newblock In {\em ICCV}, 2017.

\bibitem{GaoYSCN17}
Jiyang Gao, Zhenheng Yang, Chen Sun, Kan Chen, and Ram Nevatia.
\newblock {TURN} {TAP:} temporal unit regression network for temporal action
  proposals.
\newblock {\em ICCV}, 2017.

\bibitem{ge2019mac}
Runzhou Ge, Jiyang Gao, Kan Chen, and Ram Nevatia.
\newblock Mac: Mining activity concepts for language-based temporal
  localization.
\newblock In {\em WACV}, 2019.

\bibitem{gerhardt2013culinary}
Cornelia Gerhardt, Maximiliane Frobenius, and Susanne Ley.
\newblock {\em Culinary Linguistics}.
\newblock {John Benjamins Publishing}, 2013.

\bibitem{ghosh2019excl}
Soham Ghosh, Anuva Agarwal, Zarana Parekh, and Alexander Hauptmann.
\newblock Excl: Extractive clip localization using natural language
  descriptions.
\newblock {\em arXiv preprint arXiv:1904.02755}, 2019.

\bibitem{girdharVideoActionTransformer2019}
Rohit Girdhar, Joao Carreira, Carl Doersch, and Andrew Zisserman.
\newblock Video {{Action Transformer Network}}.
\newblock In {\em CVPR}, pages 244--253, 2019.

\bibitem{guAVAVideoDataset2018}
Chunhui Gu, Chen Sun, David~A. Ross, Carl Vondrick, Caroline Pantofaru, Yeqing
  Li, Sudheendra Vijayanarasimhan, George Toderici, Susanna Ricco, Rahul
  Sukthankar, Cordelia Schmid, and Jitendra Malik.
\newblock {{AVA}}: {{A Video Dataset}} of {{Spatio}}-{{Temporally Localized
  Atomic Visual Actions}}.
\newblock In {\em Proceedings of the {{IEEE Conference}} on {{Computer Vision}}
  and {{Pattern Recognition}}}, pages 6047--6056, 2018.

\bibitem{hahn2019tripping}
Meera Hahn, Asim Kadav, James~M Rehg, and Hans~Peter Graf.
\newblock Tripping through time: Efficient localization of activities in
  videos.
\newblock {\em arXiv preprint arXiv:1904.09936}, 2019.

\bibitem{He_2016_CVPR}
Kaiming He, Xiangyu Zhang, Shaoqing Ren, and Jian Sun.
\newblock Deep residual learning for image recognition.
\newblock In {\em The IEEE Conference on Computer Vision and Pattern
  Recognition (CVPR)}, June 2016.

\bibitem{Hendricks_2017_ICCV}
Lisa~Anne Hendricks, Oliver Wang, Eli Shechtman, Josef Sivic, Trevor Darrell,
  and Bryan Russell.
\newblock Localizing moments in video with natural language.
\newblock In {\em ICCV}, 2017.

\bibitem{huLanguageConditionedGraphNetworks2019}
Ronghang Hu, Anna Rohrbach, Trevor Darrell, and Kate Saenko.
\newblock Language-{{Conditioned Graph Networks}} for {{Relational Reasoning}}.
\newblock In {\em Proceedings of the {{IEEE International Conference}} on
  {{Computer Vision}}}, pages 10294--10303, 2019.

\bibitem{ioffe2015batch}
Sergey Ioffe and Christian Szegedy.
\newblock Batch normalization: Accelerating deep network training by reducing
  internal covariate shift.
\newblock {\em arXiv preprint arXiv:1502.03167}, 2015.

\bibitem{jainStructuralRNNDeepLearning2016}
Ashesh Jain, Amir~R. Zamir, Silvio Savarese, and Ashutosh Saxena.
\newblock Structural-{{RNN}}: {{Deep Learning}} on {{Spatio}}-{{Temporal
  Graphs}}.
\newblock In {\em Proceedings of the {{IEEE Conference}} on {{Computer Vision}}
  and {{Pattern Recognition}}}, pages 5308--5317, 2016.

\bibitem{jiang_sltfnet_2019}
Bin Jiang, Xin Huang, Chao Yang, and Junsong Yuan.
\newblock {{SLTFNet}}: {{A}} spatial and language-temporal tensor fusion
  network for video moment retrieval.
\newblock {\em Information Processing \& Management}, 56(6):102104, Nov. 2019.

\bibitem{jiang_IJMIR_2013}
Yu-Gang Jiang, Subhabrata Bhattacharya, Shih-Fu Chang, and Mubarak Shah.
\newblock High-level event recognition in unconstrained videos.
\newblock {\em International journal of multimedia information retrieval},
  2(2):73--101, 2013.

\bibitem{kingma2014method}
Diederik Kingma and Jimmy Ba.
\newblock Adam: A method for stochastic optimization, 2014.
\newblock cite arxiv:1412.6980Comment: Published as a conference paper at the
  3rd International Conference for Learning Representations, San Diego, 2015.

\bibitem{kingma_adam}
Diederik~P. Kingma and Jimmy Ba.
\newblock Adam: {A} method for stochastic optimization.
\newblock {\em CoRR}, 2014.

\bibitem{Krishna_2017_ICCV}
Ranjay Krishna, Kenji Hata, Frederic Ren, Li Fei-Fei, and Juan~Carlos Niebles.
\newblock Dense-captioning events in videos.
\newblock In {\em ICCV}, 2017.

\bibitem{krishnavisualgenome}
Ranjay Krishna, Yuke Zhu, Oliver Groth, Justin Johnson, Kenji Hata, Joshua
  Kravitz, Stephanie Chen, Yannis Kalantidis, Li-Jia Li, David~A Shamma,
  Michael Bernstein, and Li Fei-Fei.
\newblock Visual genome: Connecting language and vision using crowdsourced
  dense image annotations.
\newblock 2016.

\bibitem{kschischang2001factor}
Frank~R Kschischang, Brendan~J Frey, and H-A Loeliger.
\newblock Factor graphs and the sum-product algorithm.
\newblock {\em IEEE Transactions on information theory}, 47(2):498--519, 2001.

\bibitem{linStyleVariationCooking}
Jing Lin, Chris Mellish, and Ehud Reiter.
\newblock Style {{Variation}} in {{Cooking Recipes}}.
\newblock page~5.

\bibitem{lin_single_2017}
Tianwei Lin, Xu Zhao, and Zheng Shou.
\newblock Single {Shot} {Temporal} {Action} {Detection}.
\newblock In {\em Proceedings of the 25th {ACM} {International} {Conference} on
  {Multimedia}}, {MM} '17, pages 988--996, New York, NY, USA, 2017. ACM.
\newblock event-place: Mountain View, California, USA.

\bibitem{liu2018attentive}
Meng Liu, Xiang Wang, Liqiang Nie, Xiangnan He, Baoquan Chen, and Tat-Seng
  Chua.
\newblock Attentive moment retrieval in videos.
\newblock In {\em The 41st International ACM SIGIR Conference on Research \&
  Development in Information Retrieval}, pages 15--24. ACM, 2018.

\bibitem{Liu_ICRA_2018}
Yuxuan Liu, Abhishek Gupta, Pieter Abbeel, and Sergey Levine.
\newblock Imitation from observation: Learning to imitate behaviors from raw
  video via context translation.
\newblock 2019.

\bibitem{munLocalGlobalVideoTextInteractions2020}
Jonghwan Mun, Minsu Cho, and Bohyung Han.
\newblock Local-{{Global Video}}-{{Text Interactions}} for {{Temporal
  Grounding}}.
\newblock {\em arXiv:2004.07514 [cs]}, Apr. 2020.

\bibitem{paszke2019pytorch}
Adam Paszke, Sam Gross, Francisco Massa, Adam Lerer, James Bradbury, Gregory
  Chanan, Trevor Killeen, Zeming Lin, Natalia Gimelshein, Luca Antiga, Alban
  Desmaison, Andreas Köpf, Edward Yang, Zach DeVito, Martin Raison, Alykhan
  Tejani, Sasank Chilamkurthy, Benoit Steiner, Lu Fang, Junjie Bai, and Soumith
  Chintala.
\newblock Pytorch: An imperative style, high-performance deep learning library,
  2019.
\newblock cite arxiv:1912.01703Comment: 12 pages, 3 figures, NeurIPS 2019.

\bibitem{pech2000diatom}
Jos{\'e}~Luis Pech-Pacheco, Gabriel Crist{\'o}bal, Jes{\'u}s Chamorro-Martinez,
  and Joaqu{\'\i}n Fern{\'a}ndez-Valdivia.
\newblock Diatom autofocusing in brightfield microscopy: a comparative study.
\newblock In {\em Proceedings 15th International Conference on Pattern
  Recognition. ICPR-2000}, volume~3, pages 314--317. IEEE, 2000.

\bibitem{pennington_2014_EMNLP_glove}
Jeffrey Pennington, Richard Socher, and Christopher~D. Manning.
\newblock Glove: Global vectors for word representation.
\newblock In {\em EMNLP}, 2014.

\bibitem{ren_NIPS_2015_faster}
Shaoqing Ren, Kaiming He, Ross Girshick, and Jian Sun.
\newblock Faster {R-CNN}: Towards real-time object detection with region
  proposal networks.
\newblock In {\em NIPS}, 2015.

\bibitem{richard2018neuralnetwork}
Alexander Richard, Hilde Kuehne, Ahsan Iqbal, and Juergen Gall.
\newblock Neuralnetwork-viterbi: A framework for weakly supervised video
  learning.
\newblock In {\em IEEE Conf. on Computer Vision and Pattern Recognition},
  volume~2, 2018.

\bibitem{rodriguez2019proposal}
Cristian Rodriguez-Opazo, Edison Marrese-Taylor, Fatemeh~Sadat Saleh, Hongdong
  Li, and Stephen Gould.
\newblock Proposal-free temporal moment localization of a natural-language
  query in video using guided attention.
\newblock {\em WACV}, 2020.

\bibitem{tacos}
Anna Rohrbach, Marcus Rohrbach, Wei Qiu, Annemarie Friedrich, Manfred Pinkal,
  and Bernt Schiele.
\newblock Coherent multi-sentence video description with variable level of
  detail.
\newblock In Xiaoyi Jiang, Joachim Hornegger, and Reinhard Koch, editors, {\em
  Pattern Recognition}, 2014.

\bibitem{rohrbach2012script}
Marcus Rohrbach, Michaela Regneri, Mykhaylo Andriluka, Sikandar Amin, Manfred
  Pinkal, and Bernt Schiele.
\newblock Script data for attribute-based recognition of composite activities.
\newblock In {\em European Conference on Computer Vision}, pages 144--157.
  Springer, 2012.

\bibitem{Shou_2016_CVPR}
Zheng Shou, Dongang Wang, and Shih-Fu Chang.
\newblock Temporal action localization in untrimmed videos via multi-stage
  cnns.
\newblock In {\em The IEEE Conference on Computer Vision and Pattern
  Recognition (CVPR)}, June 2016.

\bibitem{sigurdsson_ICCV_2017}
Gunnar~A Sigurdsson, Olga Russakovsky, and Abhinav Gupta.
\newblock What actions are needed for understanding human actions in videos?
\newblock In {\em ICCV}, 2017.

\bibitem{sigurdsson2017actions}
Gunnar~A Sigurdsson, Olga Russakovsky, and Abhinav Gupta.
\newblock What actions are needed for understanding human actions in videos?
\newblock In {\em Proceedings of the IEEE International Conference on Computer
  Vision}, pages 2137--2146, 2017.

\bibitem{sigurdsson2016hollywood}
Gunnar~A. Sigurdsson, G{\"u}l Varol, Xiaolong Wang, Ali Farhadi, Ivan Laptev,
  and Abhinav Gupta.
\newblock Hollywood in homes: Crowdsourcing data collection for activity
  understanding.
\newblock In {\em European Conference on Computer Vision}, 2016.

\bibitem{szegedy2015going}
Christian Szegedy, Wei Liu, Yangqing Jia, Pierre Sermanet, Scott Reed, Dragomir
  Anguelov, Dumitru Erhan, Vincent Vanhoucke, and Andrew Rabinovich.
\newblock Going deeper with convolutions.
\newblock In {\em Proceedings of the IEEE conference on computer vision and
  pattern recognition}, 2015.

\bibitem{vaswani2017attention}
Ashish Vaswani, Noam Shazeer, Niki Parmar, Jakob Uszkoreit, Llion Jones,
  Aidan~N Gomez, {\L}ukasz Kaiser, and Illia Polosukhin.
\newblock Attention is all you need.
\newblock In {\em Advances in neural information processing systems}, pages
  5998--6008, 2017.

\bibitem{wang2019temporally}
Jingwen Wang, Lin Ma, and Wenhao Jiang.
\newblock Temporally grounding language queries in videos by contextual
  boundary-aware prediction.
\newblock {\em AAAI}, 2020.

\bibitem{xu2019multilevel}
Huijuan Xu, Kun He, L Sigal, S Sclaroff, and K Saenko.
\newblock Multilevel language and vision integration for text-to-clip
  retrieval.
\newblock In {\em AAAI}, 2019.

\bibitem{yuMAttNetModularAttention2018}
Licheng Yu, Zhe Lin, Xiaohui Shen, Jimei Yang, Xin Lu, Mohit Bansal, and
  Tamara~L. Berg.
\newblock {{MAttNet}}: {{Modular Attention Network}} for {{Referring Expression
  Comprehension}}.
\newblock In {\em Proceedings of the {{IEEE Conference}} on {{Computer Vision}}
  and {{Pattern Recognition}}}, pages 1307--1315, 2018.

\bibitem{yuan2018find}
Yitian Yuan, Tao Mei, and Wenwu Zhu.
\newblock To find where you talk: Temporal sentence localization in video with
  attention based location regression.
\newblock {\em AAAI}, 2019.

\bibitem{zengGraphConvolutionalNetworks2019}
Runhao Zeng, Wenbing Huang, Mingkui Tan, Yu Rong, Peilin Zhao, Junzhou Huang,
  and Chuang Gan.
\newblock Graph {{Convolutional Networks}} for {{Temporal Action
  Localization}}.
\newblock In {\em Proceedings of the {{IEEE International Conference}} on
  {{Computer Vision}}}, pages 7094--7103, 2019.

\bibitem{zengDenseRegressionNetwork2020}
Runhao Zeng, Haoming Xu, Wenbing Huang, Peihao Chen, Mingkui Tan, and Chuang
  Gan.
\newblock Dense {{Regression Network}} for {{Video Grounding}}.
\newblock {\em arXiv:2004.03545 [cs]}, Apr. 2020.

\bibitem{zhang2018man}
Da Zhang, Xiyang Dai, Xin Wang, Yuan-Fang Wang, and Larry~S Davis.
\newblock Man: Moment alignment network for natural language moment retrieval
  via iterative graph adjustment.
\newblock {\em CVPR}, 2019.

\bibitem{zhangWhereDoesIt2020}
Zhu Zhang, Zhou Zhao, Yang Zhao, Qi Wang, Huasheng Liu, and Lianli Gao.
\newblock Where {{Does It Exist}}: {{Spatio}}-{{Temporal Video Grounding}} for
  {{Multi}}-{{Form Sentences}}.
\newblock {\em arXiv:2001.06891 [cs]}, Jan. 2020.

\bibitem{ZhLoCoBMVC18}
Luowei Zhou, Nathan Louis, and Jason~J Corso.
\newblock Weakly-supervised video object grounding from text by loss weighting
  and object interaction.
\newblock In {\em British Machine Vision Conference}, 2018.

\bibitem{ZhXuCoCVPR18}
Luowei Zhou, Chenliang Xu, and Jason~J Corso.
\newblock Towards automatic learning of procedures from web instructional
  videos.
\newblock In {\em AAAI Conference on Artificial Intelligence}, pages
  7590--7598, 2018.

\end{thebibliography}
}

\appendix
\section*{Appendices}

\section{Datasets}

\textbf{Charades-STA}: built upon the Charades dataset \cite{sigurdsson2016hollywood} which provides time-based annotations using a pre-defined set of activity classes, and general video descriptions. In \cite{Gao_2017_ICCV}, the sentences describing the video were semi-automatically decomposed into smaller chunks and aligned with the activity classes, which were later verified by human annotators. As a result of this process, the original class-based activity annotations are effectively associated to their natural language descriptions, totalling 13,898 pairs. We use the predefined train and test sets, containing 12,408 and 3,720 moment-query pairs respectively. Videos are 31 seconds long on average, with 2.4 moments on average, each being 8.2 seconds long on average. In our ablation studies we randomly split the train set in 80\% for training and 20\% for evaluation of the experiments.

\textbf{ActivityNet Captions}: Introduced by \cite{Krishna_2017_ICCV}, this dataset which was originally constructed for dense video captioning, consists of 20k YouTube videos with an average length of 120 seconds. The videos contain 3.65 temporally localized time intervals and sentence descriptions on average, where the average length of the descriptions is 13.48 words. Following the previous methods, we report the performance  of  our algorithm on the combined two validation set.
\bigskip
\bigskip

\textbf{MPII TACoS}: built on top of the MPII Compositive dataset \cite{rohrbach2012script}, it consists of videos of cooking activities with detailed temporally-aligned text descriptions. There are 18,818 pairs of sentence and video clips in total  with the average video length being 5 minutes. A significant feature of this dataset is that due to the atomic nature of many of the descriptions ---e.g. ``takes out the knife`` and ``chops the onion''--- the associated video moments only span over a few seconds, with 8.4\% of them being less than 1.6 seconds long. This makes this dataset specially challenging for our task, as the relative brevity of the moments allows for a smaller margin of error. When it comes to splits, we use the same as in \cite{Gao_2017_ICCV}, consisting of 50\% for training, 25\% for validation and 25\% for testing.

\textbf{YouCookII}: consists on 2,000 long untrimmed videos from 89 cooking recipes obtained from YouTube by \cite{ZhXuCoCVPR18}. Each step for cooking these dishes was annotated with temporal boundaries and aligned with the corresponding section of the recipe. Recipes are written following the usual style of the domain \cite{linStyleVariationCooking,gerhardt2013culinary}, which includes very specific instruction-like statements with a wide degree of detail. The videos on this dataset are taped by individual persons at their houses while following the recipes using movable cameras. Similarly to TACoS, the average video length is 5.26 minutes. In terms of relevant moment segments, each video has 7.73 moments on average, with each segment being 19.63 seconds long on average. Videos have a minimum of 3 and a maximum of 16 moments. 

Table \ref{table:data_splits} below summarizes the details of the exact sizes of the train/validation/test splits for each dataset.

\begin{table}[h!]
    \footnotesize
    \centering
    \begin{tabular}{c c c c}
        \toprule
        \bf Dataset             & \bf Train   & \bf Validation & \bf Test   \\
        \midrule
        Charades-STA            & 12,408  & 3,720 & - \\
        \midrule
        ActivityNet Captions    & 37,414  & 17,502 & - \\
        \midrule
        YouCookII               & 10,337 & 3,492  & - \\
        \midrule
        TACoS                   & 10,146  & 4,589 & 4,083  \\
        \bottomrule
    \end{tabular}
    \caption{Exact sizes of the train/validation/test splits for each dataset. Test sets for Charades-STA, ActivityNet and YouCookII are withheld, therefore, the common practice is to report results on the validation set instead.}
    \label{table:data_splits}
\end{table}

\section{Ablated models}

In the following sub-sections we give details about the ablated models presented in Section 4.3.

\subsection{No Node Type}
\label{appe:ablated:nonodetype}

This ablation experiment is intended to show the importance of considering the Faster-RCNN features related to human labels as a different source of information. The experiment consists of assigning the same 15 object features extracted for each of the keyframes only to the Object node $\mathcal{O}$. In this way limit the ability of the network to only be able to find relations between objects and activity representations, but without reducing the total amount of data that is available to it. We consider this experiment is very relevant as it shows that the additional information provided by the objects detected is not the only reason to explain the performance improvements, but rather the way in which this data is used is more relevant. In fact enabling the model to obtain state-of-the-art performance in different and challenging benchmarks.

\subsection{No Language Attention}
\label{appe:ablated:nolanguageatte}

\begin{figure}[h]
    \centering
    \includegraphics{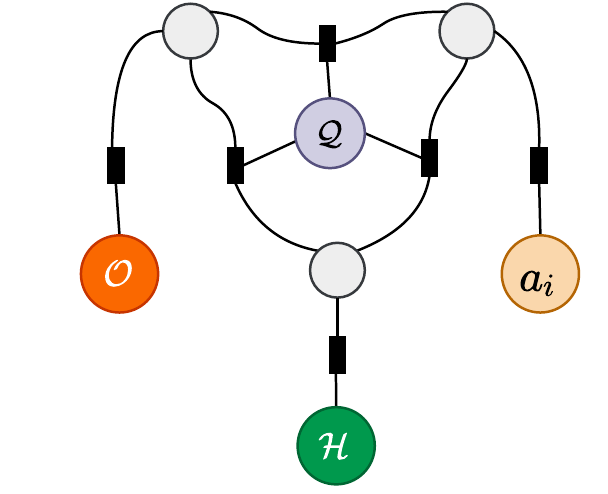}
    \caption{Spatial graph with a single query node $\mathcal{Q}$}
    \label{fig:dorimethodsupp}
\end{figure}

In this case we replace the set of linguistic nodes by a single query node $\mathcal{Q}$. It receives a high-dimensional representation (denoted by $q$) of the natural language query $Q$, as can be seen in Figure \ref{fig:dorimethodsupp}. 
This high-dimensional representation is constructed using a function $F_Q: Q \mapsto q$ that first maps each word $w_j$ for $j = 1, \ldots, m$ in the query to a semantic embedding vector $h_j \in \mathbb{R}^{d_w}$, where $d_w$ defines the hidden dimension of the word embedding. Representations for each word are then aggregated using mean pooling to get a semantically rich representation of the whole query. 

Although the query node is generic, in this work we use a bi-directional GRU \cite{cho-etal-2014-learning} on top of GLoVe word embeddings, which are pre-trained on a large collection of documents, for computing the $h_j$. Therefore, our query function $F_Q$ is parameterized by both GLoVe embedding and the GRU. 

Again we capture capture the relationship between this high-dimensional representation of the query and any observation of the nodes human $\mathcal{H}$, object $\mathcal{O}$ and activity $\mathcal{A}$, using a linear mapping function $f$ specific for each node, as follows:
\begin{align}
    \Phi_{\scriptscriptstyle\mathcal{Q,A}}^{n} &= f_{\scriptscriptstyle\mathcal{Q,A}}(q,a^{n}) \\    \Phi_{\scriptscriptstyle\mathcal{Q,O}}^{j,n} &= f_{\scriptscriptstyle\mathcal{Q,O}}(q,o^{j,n}) \\ 
    \Phi_{\scriptscriptstyle\mathcal{Q,H}}^{k,n} &= f_{\scriptscriptstyle\mathcal{Q,H}}(q,h^{k,n})
    \label{eq:relationships_single_linguistic}
\end{align}
where functions $f_{\scriptscriptstyle\mathcal{Q,A}}, f_{\scriptscriptstyle\mathcal{Q,O}}, f_{\scriptscriptstyle\mathcal{Q,H}} $ are simple linear projections. For example, in the case of the object observations we have $f_{\scriptscriptstyle\mathcal{Q,O}}(q,o^{j,n}) = W_{qo} [q;o^{j,n}] + b_{qo}$, where the subindex $qo$ denotes the dependency of the parameters of the linear function which are specific for each relation. To compute the messages that are passed between the nodes, we utilize the following functions:
\begin{align}
    \Psi^{j,n}_{\scriptscriptstyle\mathcal{H,Q,O}} &= f_{\scriptscriptstyle\mathcal{H,Q,O}}(\Phi_{\scriptscriptstyle\mathcal{Q,O}}^{j,n}, \textstyle \sum_{k=1}^K\Phi_{\scriptscriptstyle\mathcal{Q,H}}^{k,n}) \\
    \Psi^{j,n}_{\scriptscriptstyle\mathcal{A,Q,O}} &= f_{\scriptscriptstyle\mathcal{A,Q,O}}(\Phi_{\scriptscriptstyle\mathcal{Q,O}}^{j,n},\Phi_{\scriptscriptstyle\mathcal{Q,A}}^{n}) \\
    \Psi^{n}_{\scriptscriptstyle\mathcal{H,Q,A}} &= f_{\scriptscriptstyle\mathcal{H,Q,A}}(\Phi_{\scriptscriptstyle\mathcal{Q,A}}^{n}, \textstyle \sum_{k=1}^K \Phi_{\scriptscriptstyle\mathcal{Q,H}}^{k,n}) \\
    \Psi^{n}_{\scriptscriptstyle\mathcal{O,Q,A}} &= f_{\scriptscriptstyle\mathcal{O,Q,A}}(\Phi_{\scriptscriptstyle\mathcal{Q,A}}^{n}, \textstyle \sum_{j=1}^{J} \Phi_{\scriptscriptstyle\mathcal{Q,O}}^{j,n}) \\
    \Psi^{k,n}_{\scriptscriptstyle\mathcal{O,Q,H}} &= f_{\scriptscriptstyle\mathcal{O,Q,H}}(\Phi_{\scriptscriptstyle\mathcal{Q,H}}^{k,n}, \textstyle \sum_{j=1}^{J} \Phi_{\scriptscriptstyle\mathcal{Q,O}}^{j,n}) \\
    \Psi^{k,n}_{\scriptscriptstyle\mathcal{A,Q,H}} &= f_{\scriptscriptstyle\mathcal{A,Q,H}}(\Phi_{\scriptscriptstyle\mathcal{Q,H}}^{k,n},\Phi_{\scriptscriptstyle\mathcal{Q,A}}^{n})
    \label{object:messages_single_linguistic}
\end{align}
where again $f_{\scriptscriptstyle\mathcal{H,Q,O}}, f_{\scriptscriptstyle\mathcal{A,Q,O}}, f_{\scriptscriptstyle\mathcal{H,Q,A}}, f_{\scriptscriptstyle\mathcal{O,Q,A}}, f_{\scriptscriptstyle\mathcal{O,Q,H}}$ and $f_{\scriptscriptstyle\mathcal{A,Q,H}}$ are linear mappings, each receiving as input a concatenations of the corresponding features capturing. Finally, we update the representation of the human, action and object nodes based on the following formulas.
\begin{align}
    o^{j,n+1} &= \sigma(m_o(\Psi^{j,n}_{\scriptscriptstyle\mathcal{H,Q,O}} \odot \Psi^{j,n}_{\scriptscriptstyle\mathcal{A,Q,O}}) \odot o^{j,0}) \\
    a^{n+1} &= \sigma(m_a (\Psi^{n}_{\scriptscriptstyle\mathcal{H,Q,A}} \odot \Psi^{n}_{\scriptscriptstyle\mathcal{O,Q,A}}) \odot a^{0})  \\
    h^{k,n+1} &= \sigma(m_h (\Psi^{k,n}_{\scriptscriptstyle\mathcal{O,Q,H}} \odot \Psi^{k,n}_{\scriptscriptstyle\mathcal{A,Q,H}}) \odot h^{k,0}) 
    \label{other:updates_single_linguistic}
\end{align}
where $\odot$ is the element-wise product and $m_o, m_a, m_h$ are again linear functions.

\section{Language Attention}
In the following Figures \ref{fig:ln:charadessta1} and \ref{fig:ln:charadessta2}, we present a set of samples of the multihead attention to the query sentence on the Charades-STA dataset.

\begin{figure}[ht!]
    \centering
    \includegraphics[width=\linewidth]{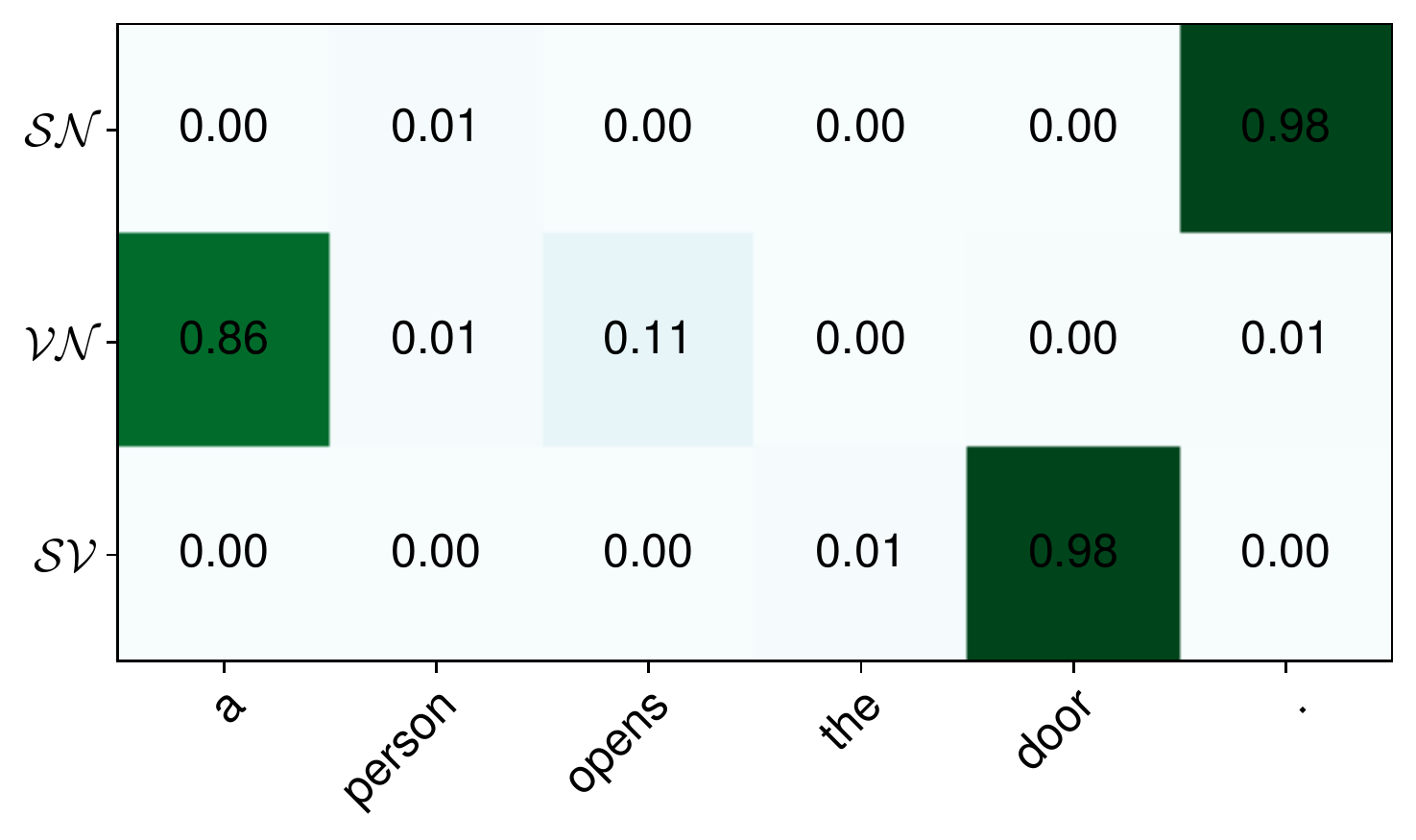}\\
    \includegraphics[width=\linewidth]{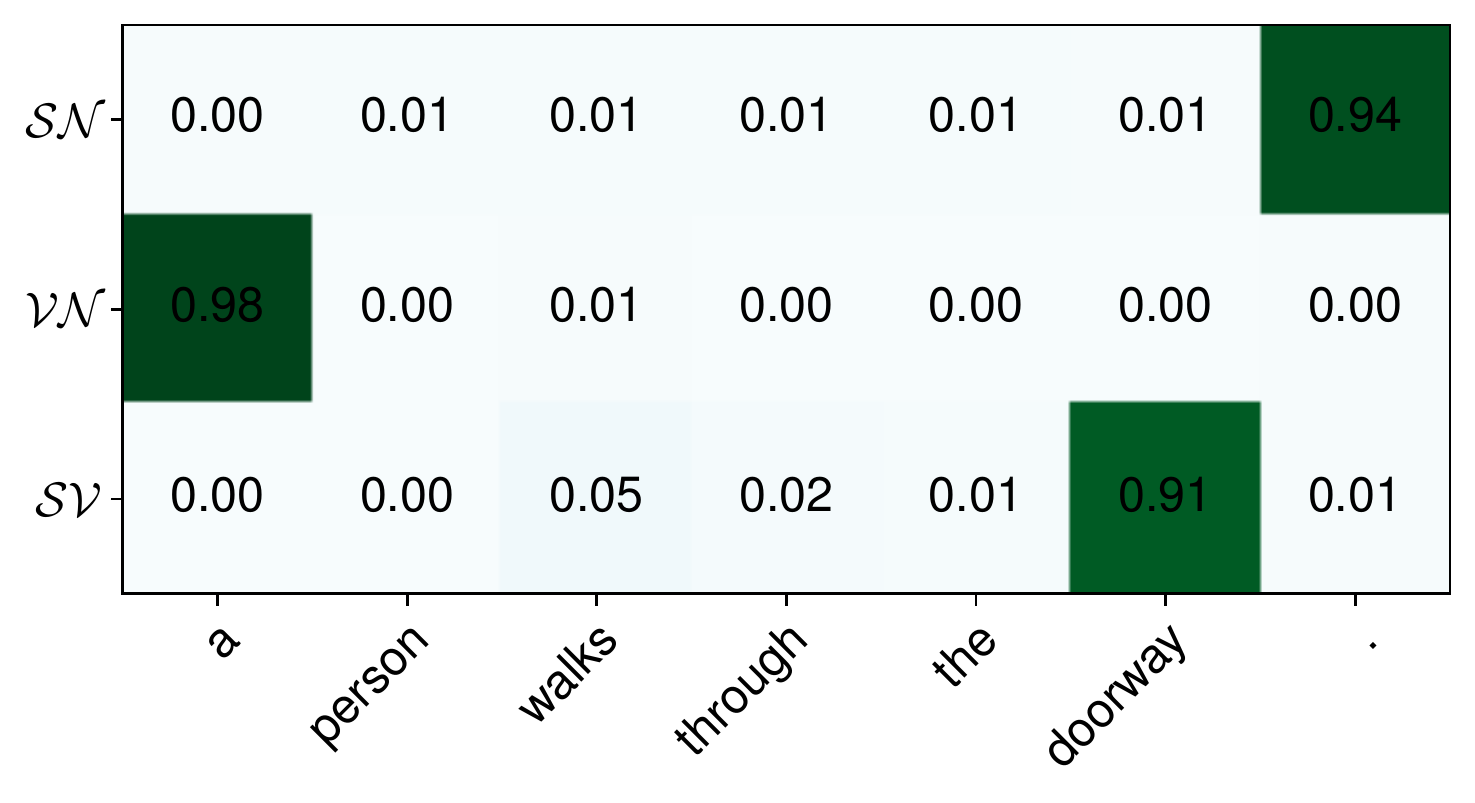}\\
    \includegraphics[width=\linewidth]{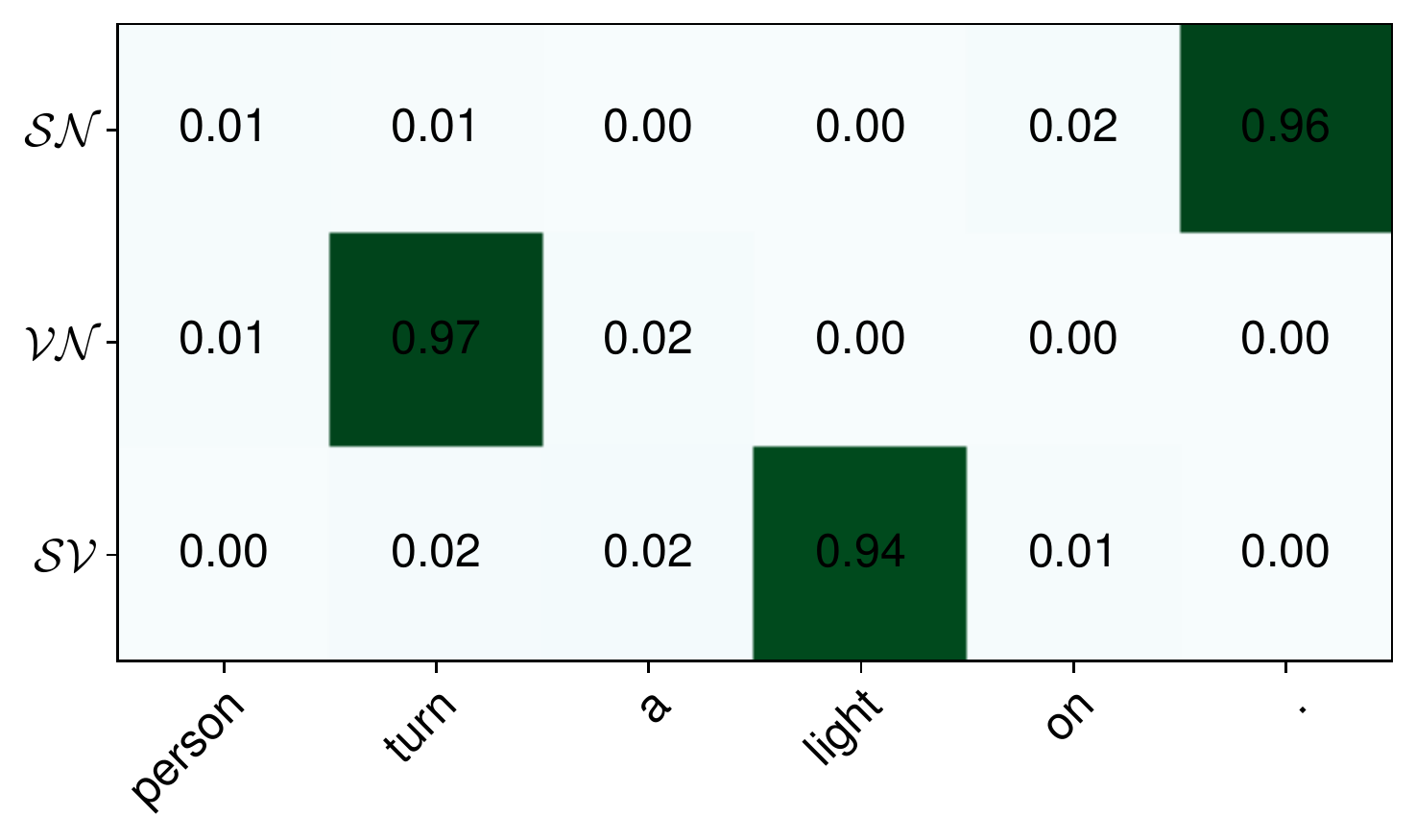}
    \caption{Linguistic nodes attentions on Charades-STA.}
    \label{fig:ln:charadessta1}
\end{figure}

\begin{figure}[h!]
    \centering
    \includegraphics[width=\linewidth]{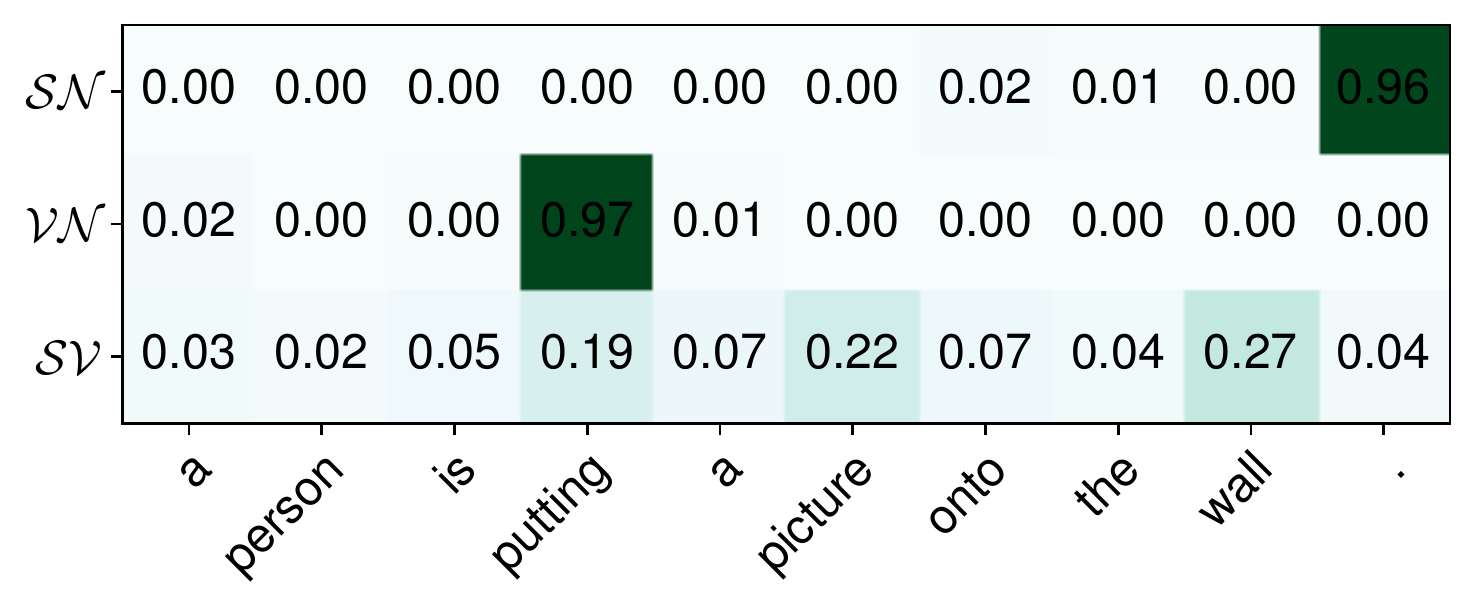}\\
    \includegraphics[width=\linewidth]{imgs/sample_a_person_is_putting_a_picture_onto_the_wall.pdf}\\
    \includegraphics[width=\linewidth]{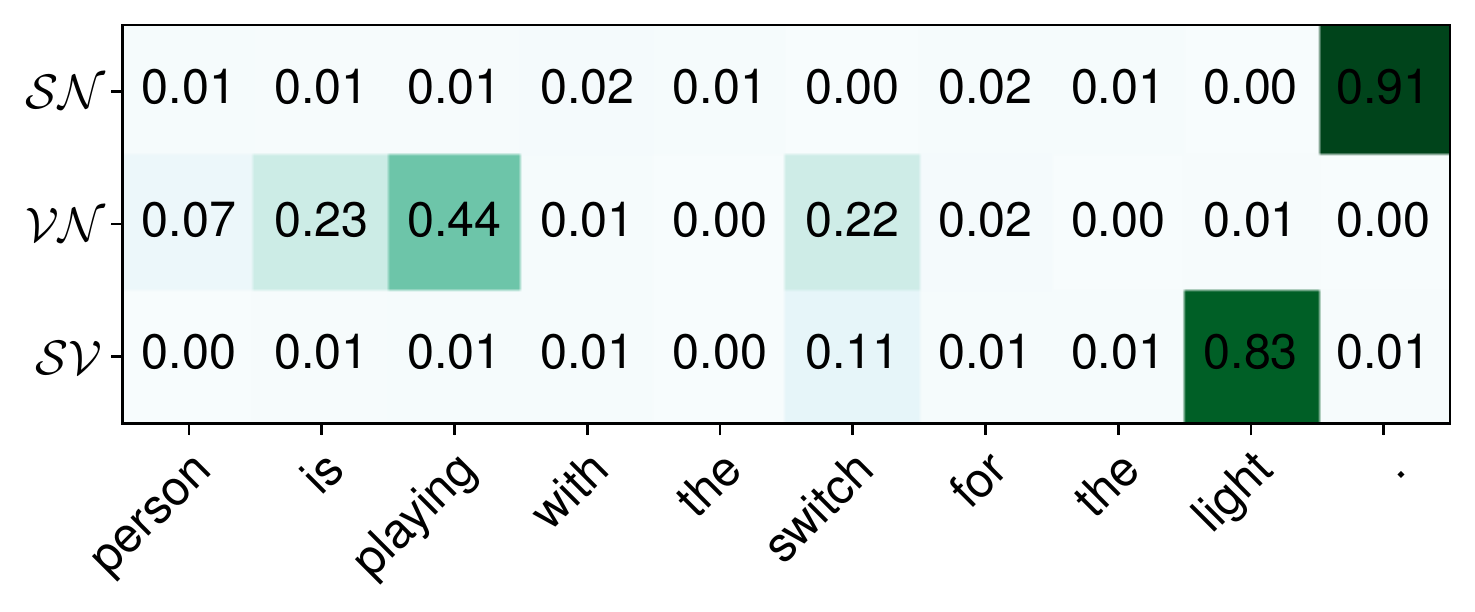}
    \caption{Linguistic nodes attentions on Charades-STA.}
    \label{fig:ln:charadessta2}
\end{figure}

\section{Examples}
\label{appe:examples}
In the following Figures, we present success and failure cases of our method on Charades-STA, YouCookII and TACoS dataset. Each visualization is showing a subsample of the keyframes inside of the prediction with their corresponding spatial observations. In green observations associated with the human node $\mathcal{H}$ and orange for the object node $\mathcal{O}$. Moreover, each visualization is presenting the ground-truth and predicted localization in seconds of the given query.
    
\subsection{Charades-STA}

\begin{figure*}[b]
    \centering
    \begin{subfigure}{\textwidth}
        \includegraphics[width=\textwidth]{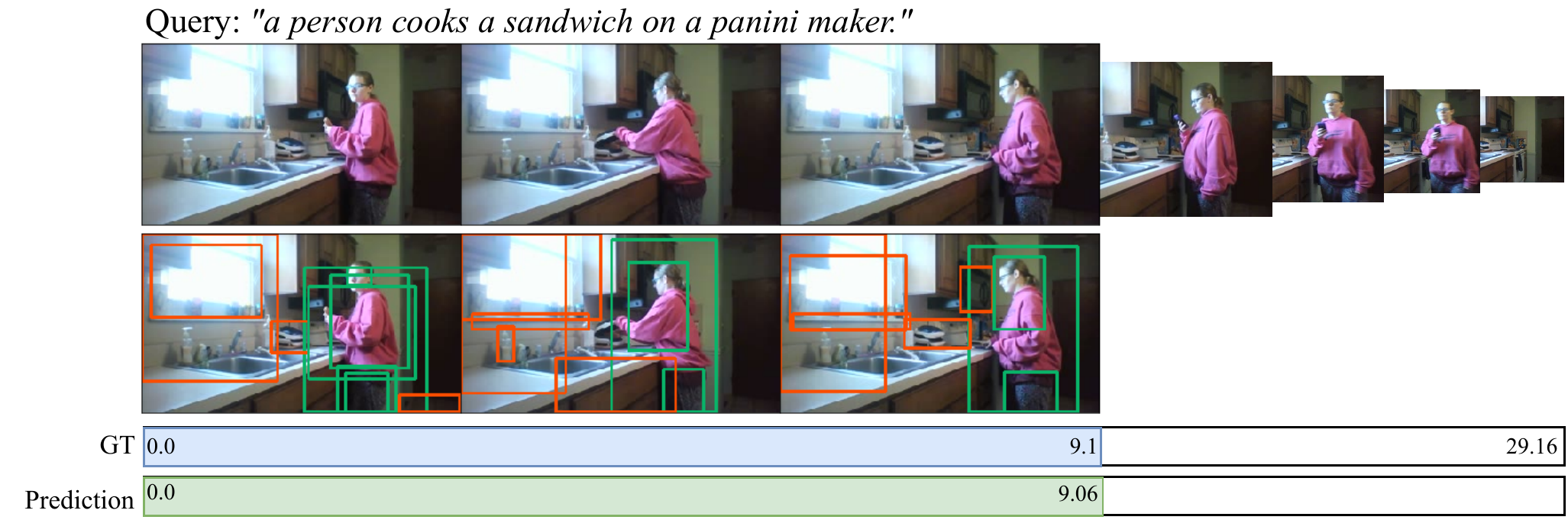}
        \caption{Example of success 1.}
        \label{fig:success:charadessta_a}
    \end{subfigure}
    \begin{subfigure}{\textwidth}
        \centering
        \includegraphics[width=\textwidth]{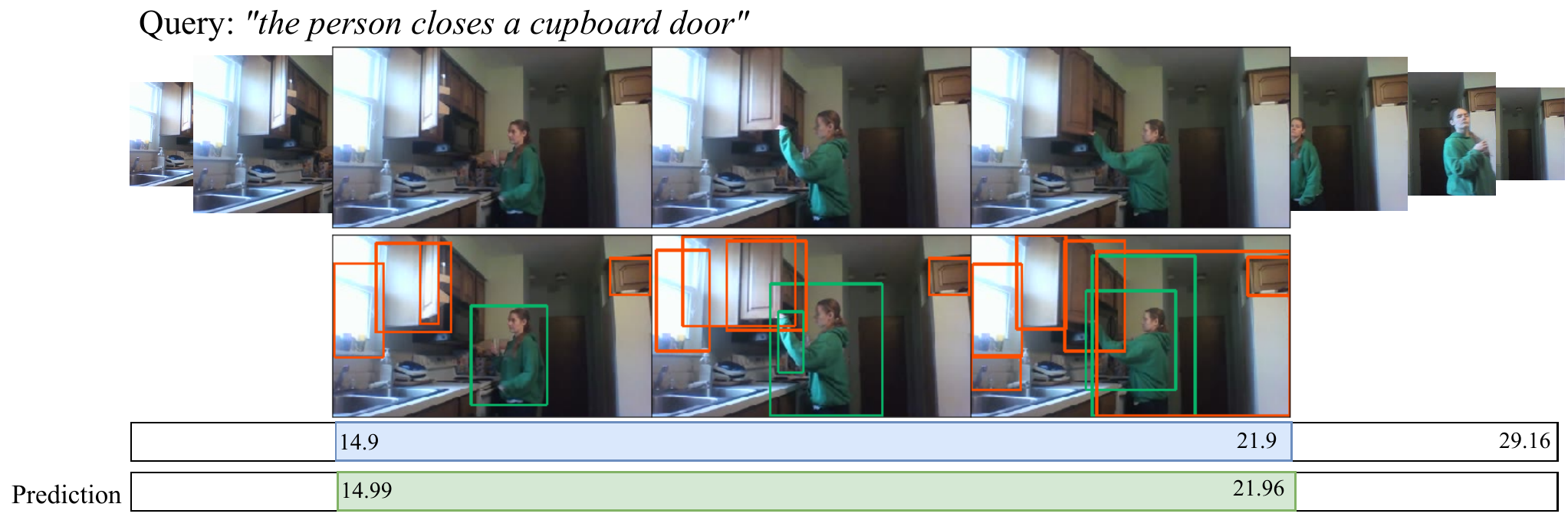}
        \caption{Example of success 2.}
        \label{fig:success:charadessta_b}
    \end{subfigure}
    \caption{Success examples of our method on Charades-STA dataset.}
    \label{fig:success:charadessta}
\end{figure*}

Success cases of our algorithm on the Charades-STA dataset can be seen in Figure \ref{fig:success:charadessta}. In Figure \ref{fig:success:charadessta_a}, given the query ``a person cooks a sandwich on a panini maker'' our method could localize the moment at a tIoU of 99.56\%. The label of the features extracted by Faster-RCNN to localize the query are {\em `bottle', `counter', `door', `drawer', `faucet', `floor', `glasses', `hair', `jacket', `jeans', 'kitchen', 'microwave', `pants', `shelf', `shirt', `sink', 'stove', 'sweater', \textbf{`toaster'}, `wall', `window', `woman'}. 

In the case of Figure \ref{fig:success:charadessta_b}, given the query ``the person closes a cupboard door.'' our method could localize the moment at a tIoU of 97.88\%. The features extracted by Faster RCNN for this query are {\em 'arm', 'building', \textbf{'cabinet'}, 'counter', 'door', 'faucet', 'hair', 'hand', 'head', 'jacket', 'kitchen', 'man', 'microwave', 'refrigerator', 'shirt', 'sink', 'sleeve', 'stove', 'sweater', 'wall', 'window', 'woman'}. 

\begin{figure*}[t]
    \begin{subfigure}{\textwidth}
        \centering
        \includegraphics[width=\textwidth]{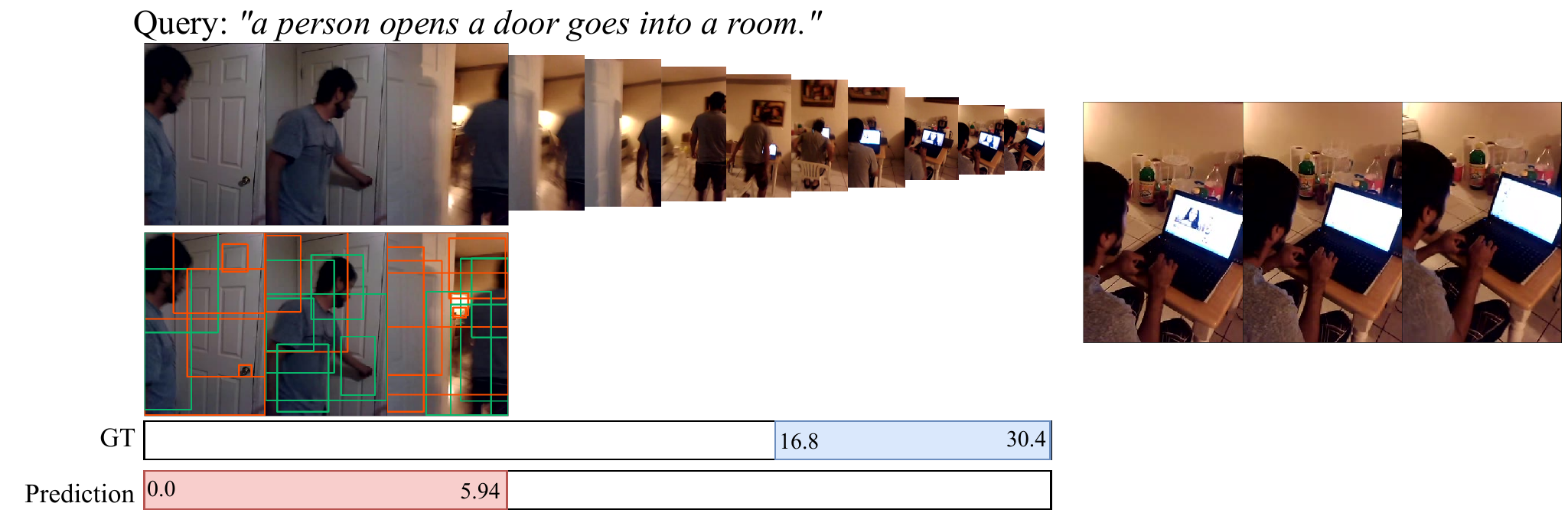}
        \caption{Example of failure 1.}
         \label{fig:fail:charadessta_a}
    \end{subfigure}
    \begin{subfigure}{\textwidth}
        \centering
        \includegraphics[width=\textwidth]{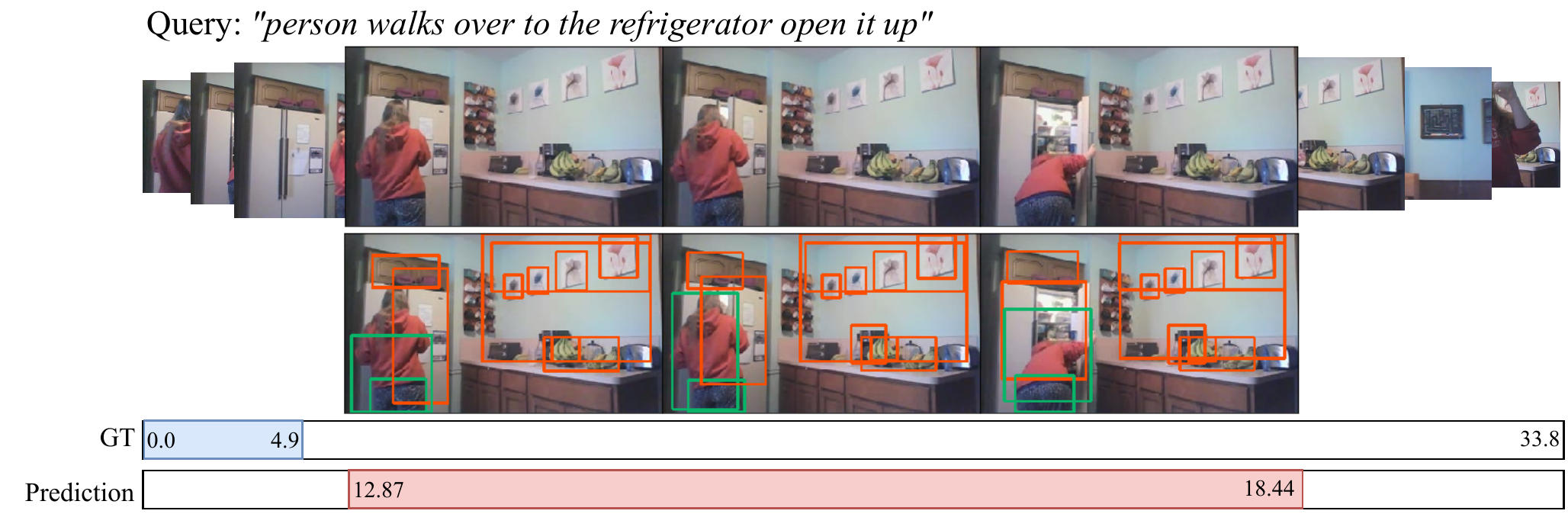}
        \caption{Example of failure 2.}
        \label{fig:fail:charadessta_b}
    \end{subfigure}
    \caption{Failure examples of our method on Charades-STA.}
    \label{fig:fail:charadessta}
\end{figure*}

Failure cases of our method are presented in Figure \ref{fig:fail:charadessta}. In the first example, given a query ``a person opens a door goes into a room." our method could detect correct spatial features, such as `door' and `knob', and the correct span of the query, according to our qualitative evaluation. However, in this case, the annotation for the query is localized incorrectly in the video. It refers to the last part of the video, where a person is using a laptop, as can be seen at the right of Figure \ref{fig:fail:charadessta_a}. In Fig. \ref{fig:fail:charadessta_b} we can see our method localizing the query ``person walks over to the refrigerator open it up'', however, the annotation is not considering that the moment is performed two times in the video.

\subsection{YouCookII}

\begin{figure*}[t]
    \begin{subfigure}{\textwidth}
        \centering
        \includegraphics[width=\textwidth]{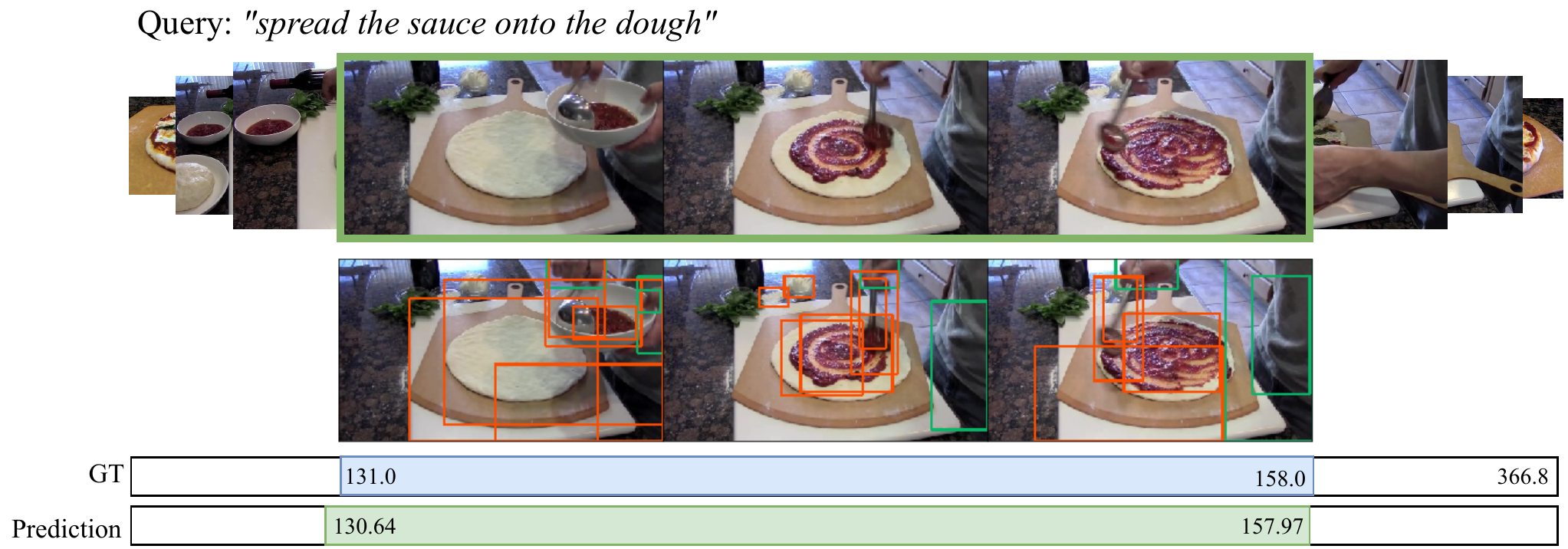}
        \caption{Success example 1.}
        \label{fig:success:youcookii_1}
    \end{subfigure}
    \begin{subfigure}{\textwidth}
        \centering
        \includegraphics[width=\textwidth]{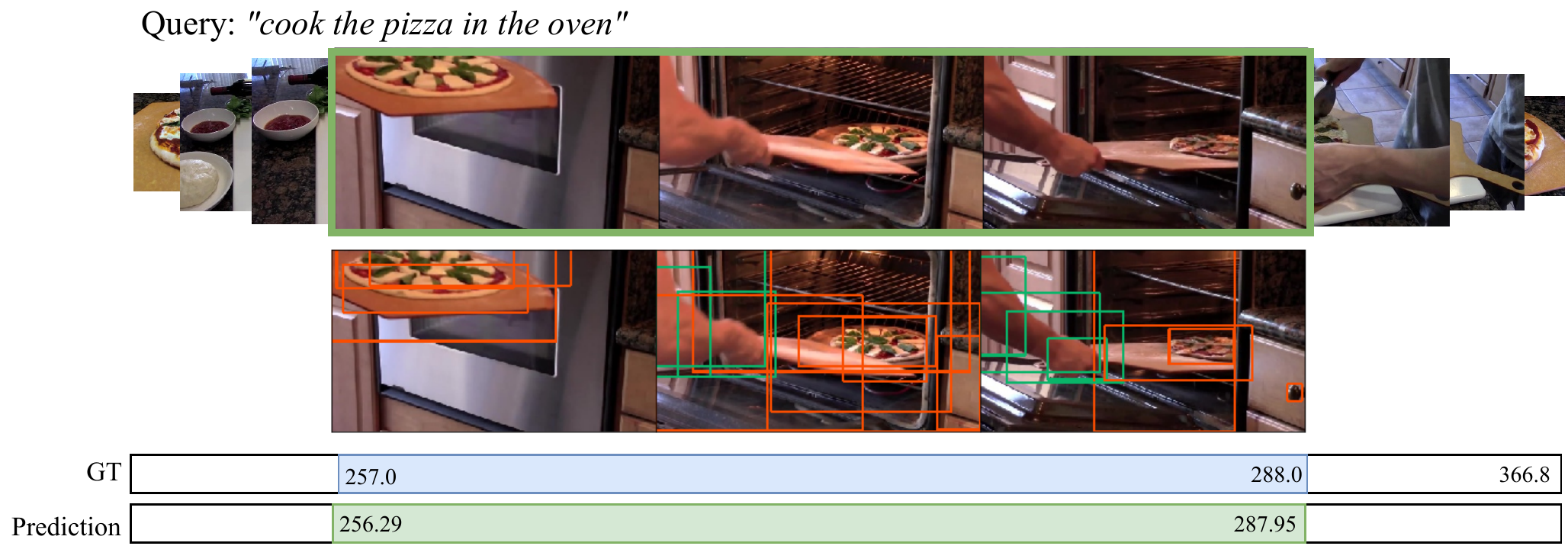}
        \caption{Success example 2.}  
        \label{fig:success:youcookii_2}
    \end{subfigure}
    \caption{Success examples of our method in the YouCookII dataset.}
    \label{fig:success:youcookii}
\end{figure*}

Although videos in YouCookII are much longer than videos in Charades-STA, our method still can get good localization performance. In Figure \ref{fig:success:youcookii_1} given the query ``spread the sauce onto the dough'' our method localize the query at a tIoU of 98.57\%. The label of the feature extracted by Faster-RCNN on this case are {\em `bacon', `bird', `board', `bottle', `bowl', `cabinet', `cake', `cherry', `chocolate', `cookie', `counter', `cutting board', `dessert', `door', `drawer', `finger', `floor', `fork', `fruit', `glass', `grape', `ground', `hand', `handle', `jeans', `ketchup', `knife', `meat', `olive', `pancakes', `pepperoni', `person', `phone', \textbf{`pizza'}, `plant', `plate', \textbf{`sauce'}, `saucer', `shirt', `sleeve', `spoon', `table', `towel', `tree', `wall'}.  

Figure \ref{fig:success:youcookii_2} shows the query ``cook the pizza in the oven'', which belong to the same video. In this case the label of the features extracted by Faster-RCNN are {\em `arm', `bar', `board', `building', `cabinet', `car', `ceiling', `cheese', `cord', `counter', `crust', `cucumber', `curtain', `door', `drawer', `fireplace', `floor', `food', `fork', `glass', `grill', `hand', `hotdog', `key', `keyboard', `kitchen', `knife', `knob', `laptop', `leaf', `leaves', `leg', `light', `man', `microwave', `mouse', \textbf{`oven'}, `oven door', `person', \textbf{`pizza'}, `plate', `pole', `rack', `roof', `room', `salad', `screen', `shadow', `sleeve', `slice', `spinach', `stove', `table', `television', `thumb', `tracks', `train', `tray', `vegetable', `vegetables', `wall', `window', `wood'} and our method could localize the query with a temporal intersection over union of 97.60\%.

\begin{figure*}[t]
    \begin{subfigure}{\textwidth}
        \centering
        \includegraphics[width=\textwidth]{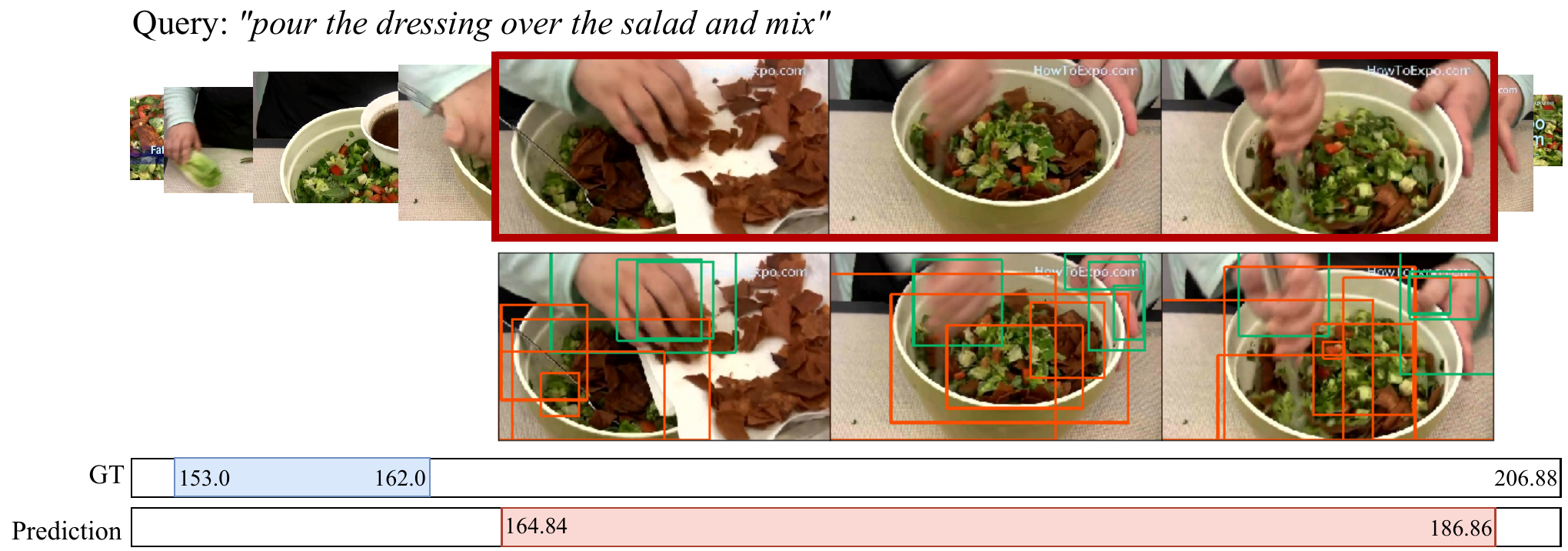}
        \caption{Failure case 1.}
        \label{fig:fail:youcookii_1}
    \end{subfigure}
    \begin{subfigure}{\textwidth}
        \centering
        \includegraphics[width=\textwidth]{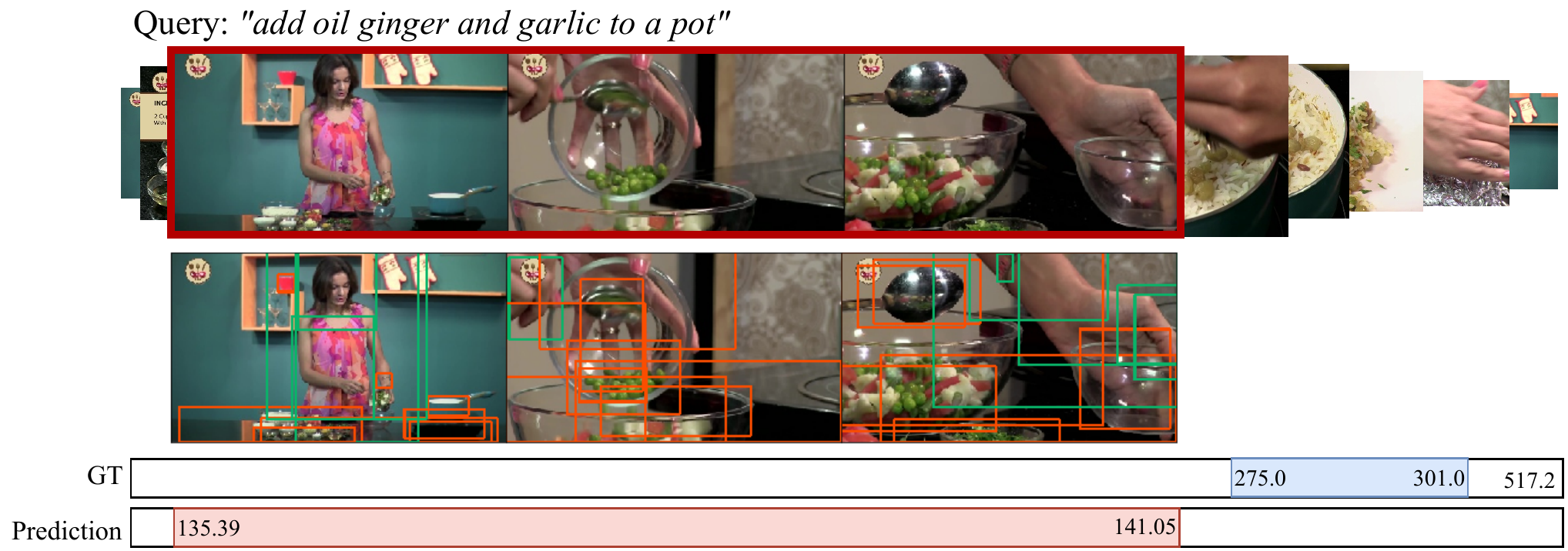}
        \caption{Failure case 2.}
        \label{fig:fail:youcookii_2}
    \end{subfigure}
    \caption{Failure cases of our method in the YouCookII dataset.}
    \label{fig:fail:youcookii}
\end{figure*}

Failure cases of our method on YouCookII dataset are presented in Figure \ref{fig:fail:youcookii}. In these cases, it is possible to see that our approach is able to recognize the activity { \em add} and {\em mix} correctly. However, the objects ``dressing, ginger and garlic'' are not detected by Faster-RCNN, probably given that the object detector has not been trained to deal with some of the kinds of objects present on this dataset. We think this naturally hinders the disambiguation capabilities of our model, specially in terms of the repetitive actions such as as adding, mixing and pouring, which are often performed throughout recipes like the one depicted in the example.

\subsection{TACoS}

\begin{figure*}[t]
    \begin{subfigure}{\textwidth}  
        \centering
        \includegraphics[width=\textwidth]{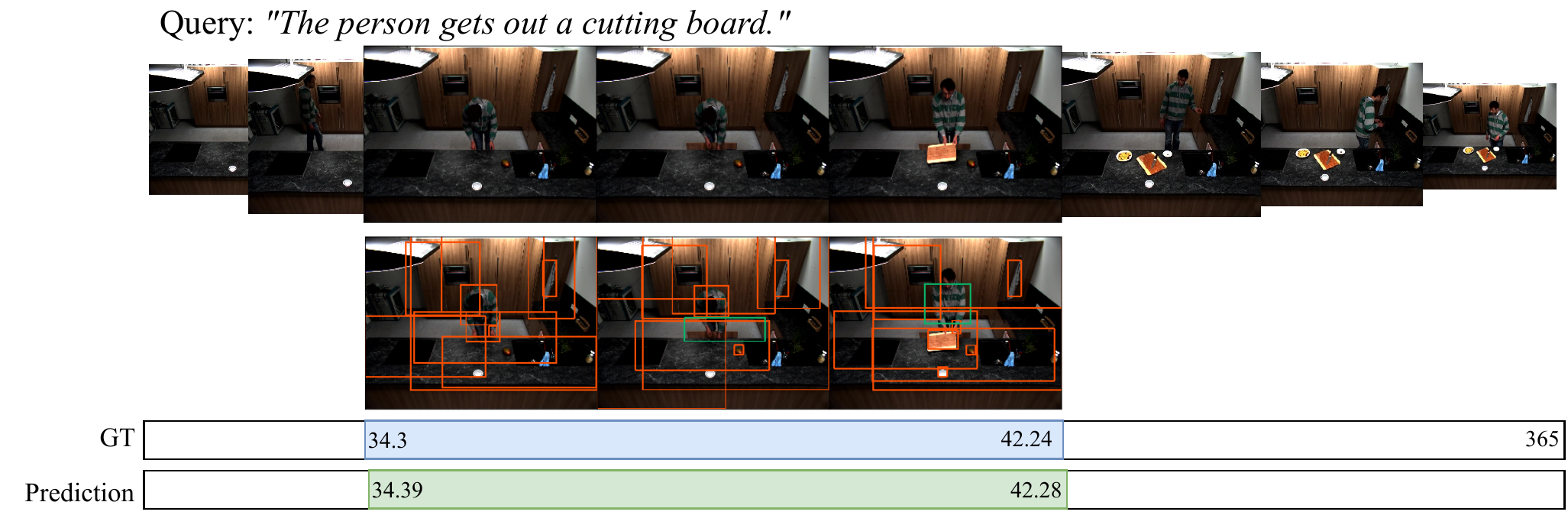}
        \caption{Success example 1.}
        \label{fig:success:tacos_1}
    \end{subfigure}
    \begin{subfigure}{\textwidth}
        \centering
        \includegraphics[width=\textwidth]{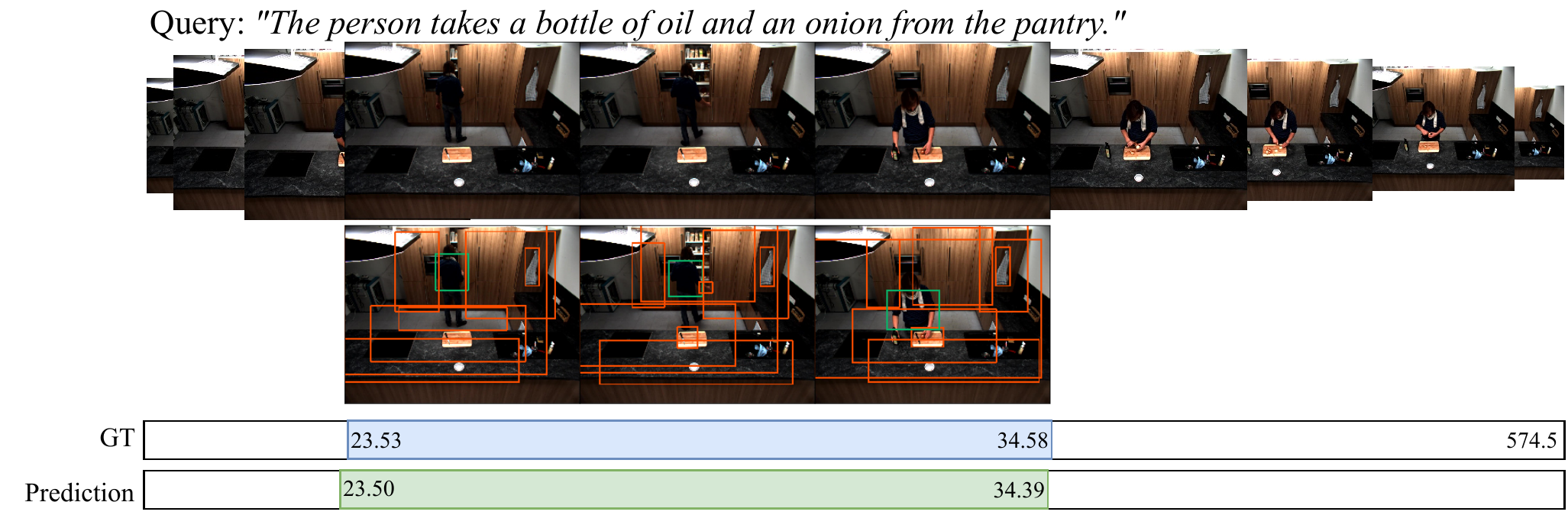}
        \caption{Success example 2.}
        \label{fig:success:tacos_2}
    \end{subfigure}
    \caption{Success examples of our method in the TACoS dataset.}
    \label{fig:success:tacos}
\end{figure*}

\begin{figure*}[t]
    \begin{subfigure}{\textwidth}
        \centering
        \includegraphics[width=\textwidth]{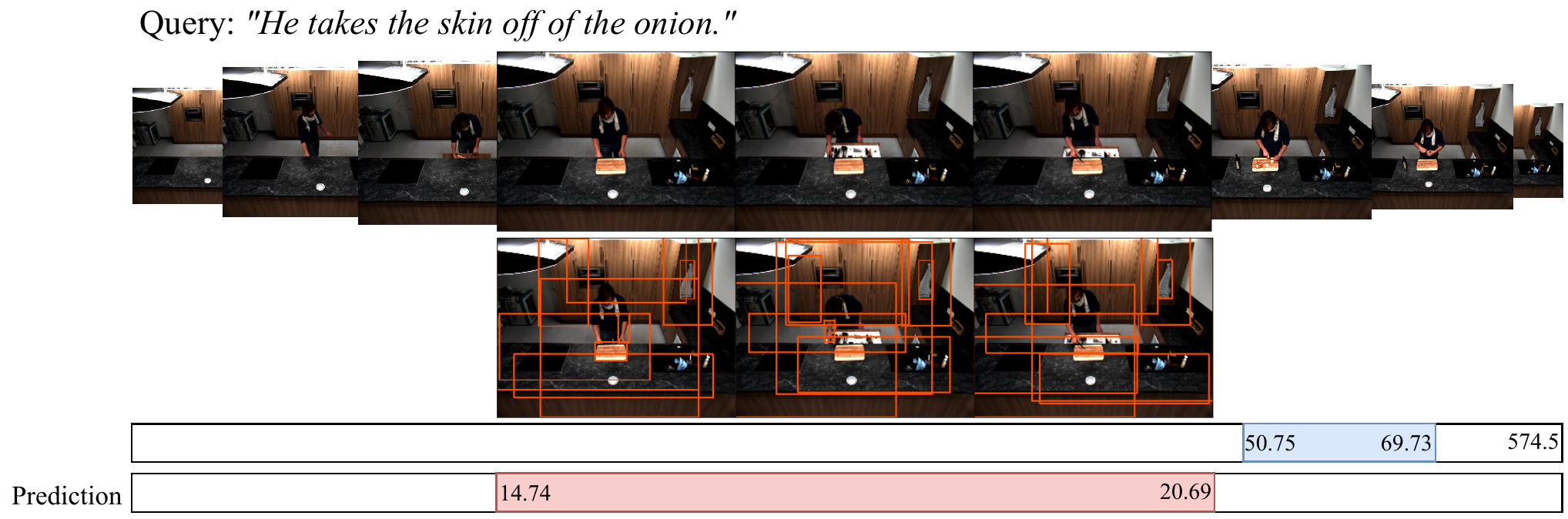}
        \caption{Failure case 1.}
        \label{fig:fail:tacos1}
    \end{subfigure}
    \begin{subfigure}{\textwidth}
        \centering
        \includegraphics[width=\textwidth]{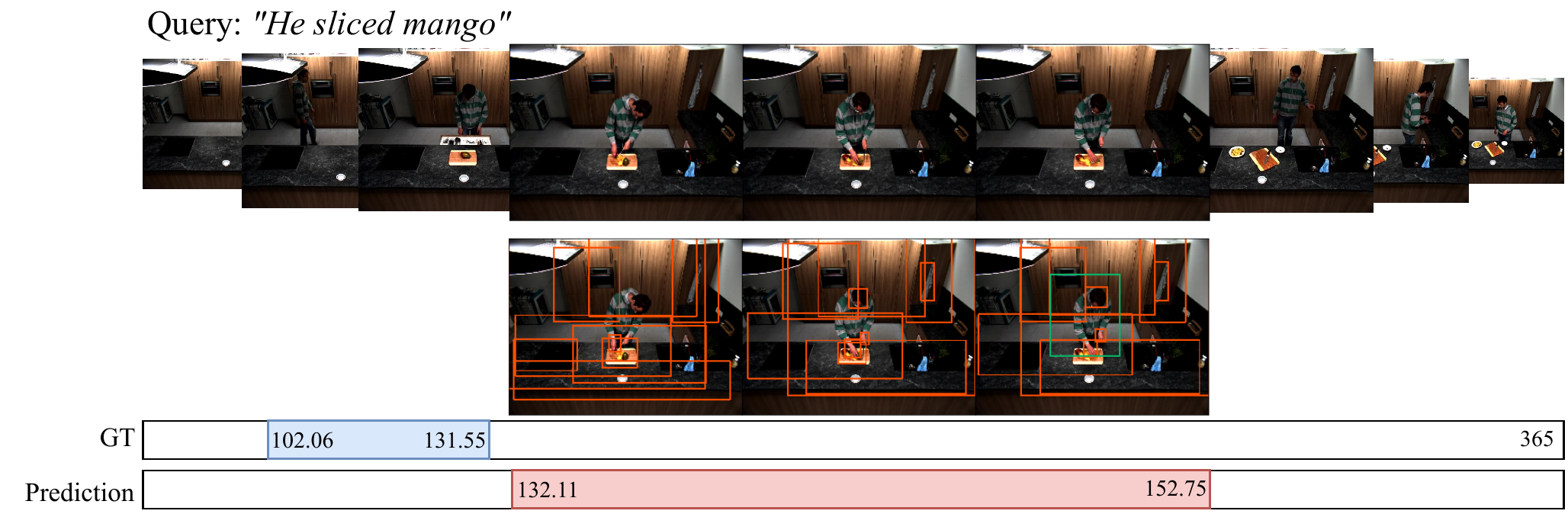}
        \caption{Failure case 2.}
        \label{fig:fail:tacos2}
    \end{subfigure}
    \caption{Failure cases of our method in the TACoS dataset.}
    \label{fig:fail:tacos}
\end{figure*}

Figures \ref{fig:success:tacos} and \ref{fig:fail:tacos} show two examples of success and failure cases on the TaCoS dataset, respectively. It is possible to see the how challenging this dataset is in general, as in the the cases where our approach fails it is in fact difficult even for us to localize the given query.

\section{Experimental Information}

Our models are implemented using PyTorch \cite{paszke2019pytorch} and are trained using the Adam \cite{kingma2014method} optimizer, with a batch size of 6. 
Experiments for different datasets were run in two different machines:

\begin{itemize}
    \item First server machine with an Intel Core i7-6850K CPU with two NVIDIA Titan Xp (Driver 430.40, CUDA 10.1) GPUs, and one NVIDIA Quadro P5000, running ArchLinux

    \item An additional server machine with an Intel Xeon 4215 CPU, with three NVIDIA RTX8000 (Driver 430.44, CUDA 10.1) GPUs, running Ubuntu 16.04 
\end{itemize}

We used PyTorch version 1.4. Our method has 10.865.155 trainable parameters. In training takes 1.56 hours per epoch in Charades-STA, 4.3 hours per epoch in TACoS, 5.4 hours per epoch in YouCookII and 6.7 hours per epoch in ActivityNet. In average our method takes 0.015 seconds to localize one query.

\end{document}